\documentclass[]{interact}

\usepackage{epstopdf}% To incorporate .eps illustrations using PDFLaTeX, etc.

\usepackage{natbib}% Citation support using natbib.sty
\bibpunct[, ]{(}{)}{;}{a}{}{,}% Citation support using natbib.sty
\renewcommand\bibfont{\fontsize{10}{12}\selectfont}% Bibliography support using natbib.sty

\theoremstyle{plain}% Theorem-like structures provided by amsthm.sty

\theoremstyle{definition}

\theoremstyle{remark}

\usepackage{adjustbox}
\usepackage{tikz}
\usepackage{caption}
\usepackage{subcaption}
\usepackage{amssymb}
\usepackage{tikz}
\usepackage{lineno}
\usepackage{fancyhdr,lipsum}
\usepackage{setspace}
\usepackage{xspace}
\usepackage{amssymb}
\usepackage{booktabs}
\usepackage{diagbox}
\usepackage{graphicx}
\usepackage{float} 
\usepackage{caption}
\usepackage{parskip}
\usepackage{amsmath}
\usepackage{multirow}
\usepackage{makecell}
\usepackage{algorithm}
\usepackage{algpseudocode}
\usepackage{amsthm}
\floatstyle{plaintop}
\restylefloat{table}
\usepackage[colorlinks,
            linkcolor=blue,
            anchorcolor=blue,
            citecolor=blue
            ]{hyperref}

\newcommand*{\eg}{e.g.\@\xspace}
\makeatletter
\newcommand*{\etc}{%
    \@ifnextchar{.}%
        {etc}%
        {etc.\@\xspace}%
}
\begin{document}

\articletype{}

\title{Learning county from pixels: corn yield prediction with attention-weighted multiple instance learning}

% \author{
% \name{A.~N. Author\textsuperscript{a}\thanks{CONTACT A.~N. Author. Email: latex.helpdesk@tandf.co.uk} and John Smith\textsuperscript{b}}
% \affil{\textsuperscript{a}Taylor \& Francis, 4 Park Square, Milton Park, Abingdon, UK; \textsuperscript{b}Institut f\"{u}r Informatik, Albert-Ludwigs-Universit\"{a}t, Freiburg, Germany}
% }

\author{
\name{Xiaoyu Wang\textsuperscript{a}, Yuchi Ma\textsuperscript{b}, Yijia Xu\textsuperscript{a}, Qunying Huang\textsuperscript{c}, Zhengwei Yang\textsuperscript{d}, Zhou Zhang\textsuperscript{a}\thanks{CONTACT Zhou Zhang Email: zzhang347@wisc.edu}}
\affil{\textsuperscript{a}Biological Systems Engineering, University of Wisconsin-Madison, Madison, 53706, WI, USA; \textsuperscript{b}Department of Earth System Science and Center on Food Security and the Environment, Stanford University, Stanford, 94305, CA, USA; \textsuperscript{c}Department of Geography, University of Wisconsin-Madison, Madison, 53706, WI, USA; \textsuperscript{d}Research and Development Division, National Agricultural Statistics Service, United States Department of Agriculture, Washington, 20250, DC, USA}
}

\maketitle

\begin{abstract}
Remote sensing has become a promising tool in yield prediction. Most prior work using satellite imagery for county-level corn yield prediction spatially aggregates all pixels within a county into a single value, potentially overlooking the detailed information and valuable insights offered by more granular data. This research studies each county at the pixel level and applies multiple instance learning to leverage the detailed information within each county. In addition, our method addresses the ``mixed pixel'' problem caused by the inconsistent resolution between feature datasets and crop masks, which may introduce noise into the model and make corn yield prediction more difficult. Specifically, the attention mechanism is utilized to automatically assign weights to different pixels, which can mitigate the influence of the ``mixed pixel'' problem. The experimental results show that the developed model outperforms four other machine learning models over the past five years in the U.S. Corn Belt and demonstrates its best performance in 2022, achieving a coefficient of determination ($R^2$) value of 0.84 and a root mean square error ($RMSE$) value of 0.83. This paper demonstrates the advantages of our approach from both spatial and temporal perspectives. Furthermore, through an in-depth study of the relationship between mixed pixels and attention, it is verified that our approach can capture important feature information while filtering out noise from mixed pixels.
\end{abstract}

This work was supported by the United States Department of Agriculture (USDA) National Institute of Food and Agriculture, Agriculture and Food Research Initiative Project under Grant 1028199.

% \begin{keywords}
% Corn; Yield prediction; Machine learning; Multiple instance learning; Attention
% \end{keywords}

% \linenumbers

\section{Introduction}
\label{Introduction}

County-level corn yield prediction in the U.S. holds significant importance due to its central role in the country's agriculture and economy \citep{rse1,rse2,rse3}. Accurate predictions enable better planning for supply chain management, price setting, and policy-making, as well as assisting farmers in the decision-making process to optimize resource use. Due to the advantages of efficiency, easy data acquisition, and wide spatial coverage, remote sensing (RS) technology has been widely applied to yield prediction \citep{rse4,rse5,rse6}. RS for yield prediction typically involves two main approaches: process-based method \citep{process1,process2,process3,process4} and learning-based method \citep{rse1,rse3}. The former utilizes mathematical models to represent and simulate the physical and biological processes that determine crop growth and development. The latter predicts yield by learning the complex and non-linear relationships between yield and features extracted from RS data. As a data-driven approach, the learning-based method can quickly process large volumes of data, leading to better generalization on new data, without the need for explicitly modeling the physical processes as in the process-based method \citep{rse7,rse8,rse9}. This study focuses on exploring the use of a learning-based method for the county-level corn yield prediction task.

Deep learning \citep{dlbook} models are currently the most popular and effective learning-based models. Many studies have investigated the use of various types of neural networks with RS data for the crop yield prediction task. One of the earliest neural networks, the Multi-Layer Perceptron (MLP) \citep{mlp}, consists of multiple layers of nodes in a directed graph and can model complex data \citep{mlp_1, mlp_2, mlp_3}. The Convolutional Neural Network (CNN) \citep{cnn} uses a mathematical operation called convolution, which is highly effective in processing images \citep{cnn_1, cnn_2, cnn_3}. For time series modeling, the Recurrent Neural Network (RNN) \citep{rnn} with its internal memory can capture information about previous steps \citep{rnn_1, rnn_2, rnn_3}. Long Short-Term Memory (LSTM) networks \citep{lstm} are good at modeling temporal sequences and capturing phenological trends over time \citep{lstm_1}. By combining CNN and LSTM, we can utilize both spatial and temporal information \citep{cnn_lstm_1}. Furthermore, ResNet \citep{resnet} employs residual learning, which helps mitigate the vanishing gradient problem in deep networks and improves feature extraction and accuracy \citep{resnet_1}. Recently, the Transformer \citep{attention} model has significantly improved the performance in various language tasks and has become the foundation of multi-modal combination \citet{transformer_0}. Various Transformer variants, like the Vision Transformer (ViT) \citep{vit} and Swin Transformer (Swin-T) \citep{swin}, have also become mainstream models in remote sensing image processing, demonstrating the potential and rapid development of deep learning for yield prediction tasks \citep{transformer_2, transformer_4}.

In county-level yield prediction, many previous studies treat a county as a unified entity, aggregating all pixels within the county to a single value (\eg mean value) \citep{county1, county2, rse7, rse3}. This method not only simplifies the model by considerably reducing the computational load but also results in accurate predictions. For example,  \citet{county2} develops a dual-branch deep learning framework for county-level winter wheat yield prediction, where each county's RS information is spatially compressed into the average value.  \citet{rse3} employs a Bayesian neural network for both county-level corn yield prediction and its corresponding predictive uncertainty. This study utilizes the median value instead of average to effectively reduce the influence of outliers on the yields \citep{county1}. However, aggregating all the pixels within the county to a single value may compromise the integrity of each county's data, potentially losing detailed information. As a result, missing crucial information could substantially reduce the model's predictive performance. Therefore, we collect all the pixels of cornfields from each county and utilize them as completely as possible to predict the county's yield, thereby enhancing the accuracy and reliability of the predictions.

% Introduce mir
To better utilize the information from pixels within each county, we employ Multiple Instance Learning (MIL) for modeling. MIL \citep{mir,surveymir} is a variant of a supervised machine learning method. In traditional supervised learning, an \emph{instance} refers to a single data point. Each instance has one corresponding label. Whereas in MIL, labels are assigned to \emph{bags} which are collections of instances, rather than to the individual instances within these bags. MIL is designed to leverage the detailed information in a bag to predict the label of the entire bag, even though individual instances inside the bag are not explicitly labeled. When predicting county-level corn yield by treating a county as an aggregation of all individual pixels, each \emph{county} can be viewed as a \emph{bag} and its comprising \emph{pixels} can be treated as \emph{instances}. To fully utilize the connection of instances in a bag, Cluster-MIL \citep{clustermir} assumes that individual instances come from a group of underlying clusters, with a bag's label being a function of one relevant cluster. The model learns the internal structure of bags that contain instances from various unknown distributions. Prime-MIL \citep{primemir} supposes there exists a primary instance within each bag, then aims to select the most important instance for prediction. In contrast, Instance-MIL (Ins-MIL) \citep{insmir} employs each instance for regression directly, without taking the bag structure into account. There are two reasons that motivate us to investigate using MIL for county-level corn yield prediction modeling. First, MIL can make use of the detailed information in the pixels within a county for yield prediction, potentially utilizing underlying patterns within a county to assist the learning process. Second, in our scenario, only pixel-level RS imagery is available, but pixel-level yield records are lacking, making end-to-end pixel-level learning and analysis difficult. MIL can train models and make predictions even without instance labels, which avoids the end-to-end pixel-level learning. 

\begin{figure}
    \captionsetup[subfigure]{labelformat=empty}
    \tikzset{inner sep=0pt}
    \setkeys{Gin}{width=0.22\textwidth}
    \centering
    
    \subfloat[\label{mysubfig1}]{%
    \tikz{\node (a) {\includegraphics[width=0.4\textwidth]{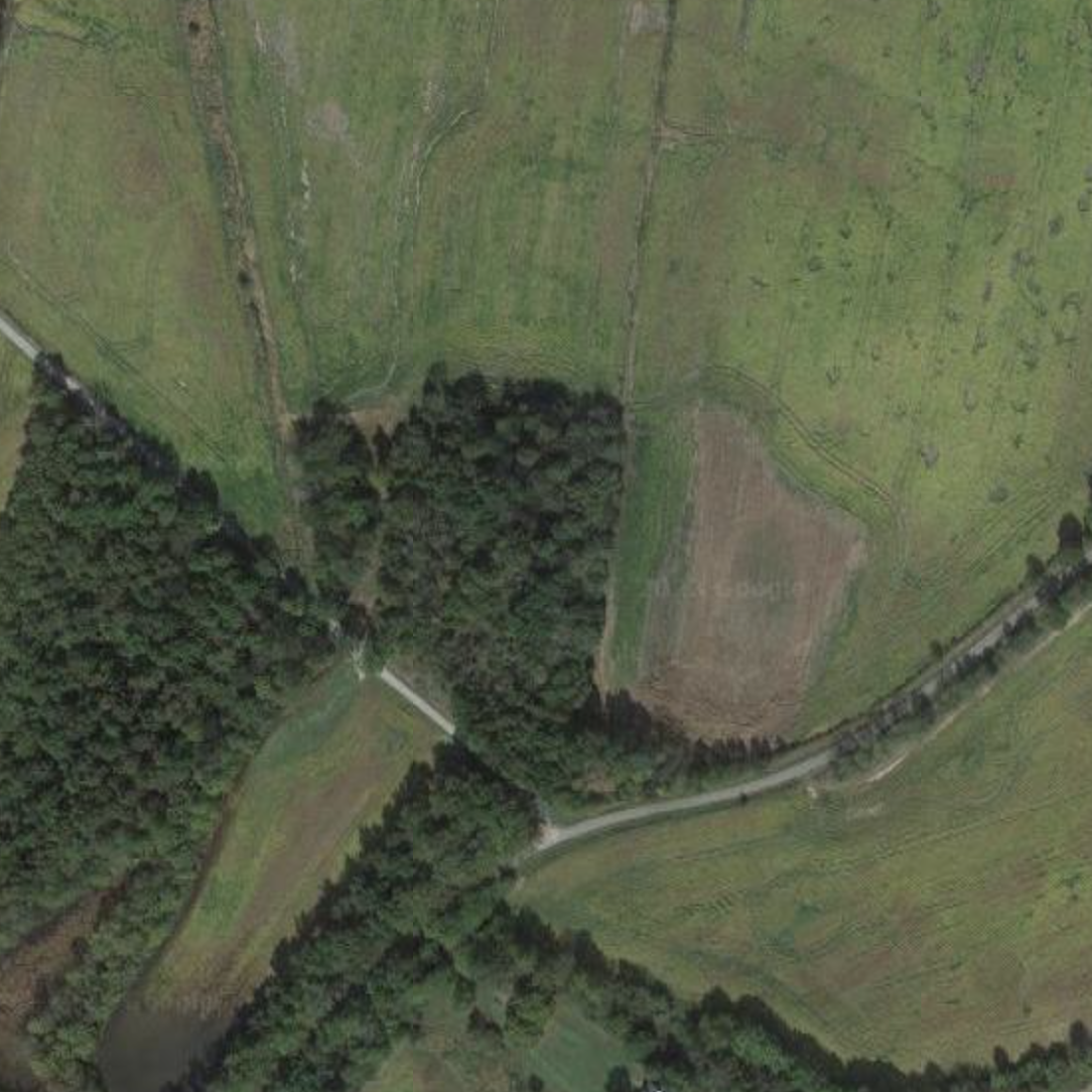}};
          \node[above, yshift=-1cm] at (a.south) {(a)};
          }}
    % \hfill
    \hspace{0.5cm}
    \subfloat[\label{mysubfig2}]{%
    \tikz{\node (a) {\includegraphics[width=0.4\textwidth]{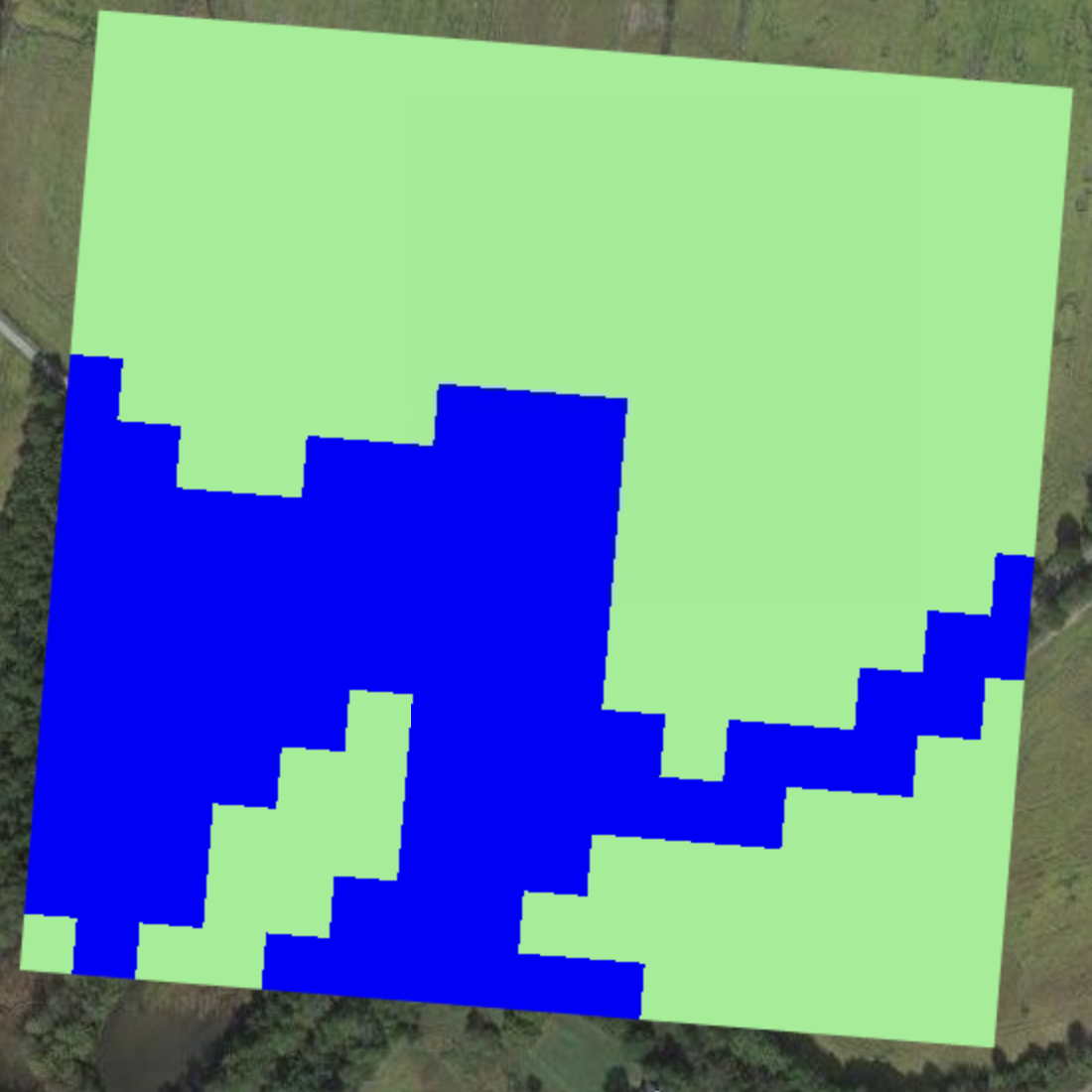}};
          \node[above, yshift=-1cm] at (a.south) {(b)};
          }}

    \caption{The map from Google Earth of (a) One $500\mathrm{m} \times 500\mathrm{m}$ MODIS pixel and (b) this $500\mathrm{m} \times 500\mathrm{m}$ MODIS pixel masked by $30\mathrm{m} \times 30\mathrm{m}$ resolution CDL mask.}
    \label{fig:mixedmodis}
\end{figure}

When employing MIL to better utilize detailed information, we study satellite imagery at the pixel level and encounter a challenge commonly referred to as the ``mixed pixel'' problem \citep{mix_1}, which arises from the inconsistent resolution between crop masks and various feature datasets. We use the MODIS dataset \citep{modis,modismcd} as a primary example to illustrate this problem. \autoref{fig:mixedmodis} (a) shows a $500\mathrm{m} \times 500\mathrm{m}$ MODIS pixel, which contains a patch of land with corn, bare soil, and weeds. \autoref{fig:mixedmodis} (b) displays the same MODIS pixel masked by the $30\mathrm{m} \times 30\mathrm{m}$ CDL crop mask \citep{cdl}, where the green portion represents land planted with corn, and the blue portion indicates unrelated land types. This large MODIS pixel is inherently impure because it encompasses multiple land types. In our research, the primary focus is on the $30\mathrm{m} \times 30\mathrm{m}$ CDL cornfield pixels segmented by the CDL mask. When we study such a CDL cornfield pixel in the green area of \autoref{fig:mixedmodis} (b), its feature information, such as the Vegetation Indices (VIs), is derived from the encompassing large MODIS pixel, which contains a mixture of different land types. Therefore, even if this small CDL pixel is entirely composed of cornfield, its information is still mixed. This introduces noise into the models we build, affecting their performance. As a result, we refer to this small CDL cornfield pixel containing mixed information as a ``mixed pixel''. In subsequent sections, any reference to a mixed pixel specifically indicates the small CDL cornfield pixels under investigation. It is worth noting that merely increasing the resolution of the crop mask does not solve the mixed pixel problem, given the inherent resolution limitations of satellite imagery. In addition, although we use the MODIS dataset as an illustration, the mixed pixel problem is not unique to satellite imagery; any feature with a resolution lower than that of the CDL mask can present this problem. To effectively address this issue, we propose using an attention mechanism \citep{attention} to rationally utilize the information within these mixed pixels.
% \clearpage

\clearpage

% \clearpage

To address the ``mixed pixel'' problem, a straightforward approach is to calculate the mixing level of these mixed pixels and then directly use this mixing level to weight each pixel. However, this method poses certain challenges and limitations. When employing various datasets with different spatial resolutions, the choice of scale for calculating mixing level needs to be considered. It's difficult to manually integrate the mixing levels from different datasets. Additionally, we also discuss the disadvantages of this method in detail in the discussion (\autoref{Discussion}). Therefore, we use an attention mechanism to automatically calculate the weights of mixed pixels. The attention mechanism \citep{attention,attmir} is a powerful tool in deep learning. It enables models to focus on specific aspects of complex input data. Essentially, it can be used to assign different weights or ``attention scores'' to different parts of the input. This weighting mechanism is analogous to the way human beings pay attention to certain details while ignoring others \citep{attentionsurvey}. To address the ``mixed pixel'' problem, we propose to use this attention mechanism to allocate higher weights to pixels predominantly composed of cropland and lower weights to those with more irrelevant land types, thereby improving the model's accuracy and interpretability. This method effectively utilizes the information from each pixel while mitigating the negative impact of noisy information in mixed pixels.

Many previous studies apply the combination of the attention mechanism and MIL. For example, \citet{attmir} introduces a MIL method that is completely driven by neural networks, which includes trainable MIL pooling with the gated attention mechanism. \citet{attmil1} incorporates self-attention into the process of MIL Pooling, merging the multi-level dependencies among different image areas with a trainable weighted average pooling mechanism. \citet{attmil2} proposes a new MIL technique for medical image analysis named triple-kernel gated attention-based MIL with contrastive learning instead of pre-training. While the above studies have explored the combination of attention and MIL extensively, this study aims to effectively utilize detailed information through MIL and subsequently resolve the mixed pixel problem with an attention mechanism, which are two progressive objectives. Furthermore, in studies on the mixed pixel problem, no one has directly applied attention to this issue \citep{mix_1, mix_2}. To the best of our knowledge, this is the first application of the attention mechanism to address the mixed pixel problem in county-level corn yield prediction.

% Summary 
This paper presents a new method to enhance county-level corn yield prediction. Previous work mostly aggregated features to the county level, thereby ignoring pixel-level information. Our method utilizes the MIL framework to extract detailed information from individual pixels within each county, thereby enhancing performance. Furthermore, to solve the ``mixed pixel'' problem, our method uses the attention mechanism, which automatically assigns weights to each pixel based on its contribution to the final prediction. In the experimental validation, we not only achieved good results using our method on data from the U.S. Corn Belt but also verified the reliability of the results using some visualization techniques. To demonstrate the superiority of our method, we compared it with traditional machine learning methods and several promising deep learning models for county-level corn yield prediction. Currently, apart from the classic MLP \citep{mlp}, CNN \citep{cnn}, and LSTM \citep{lstm}-based methods, the most popular deep learning model in remote sensing is the Transformer-based \citep{attention} model \citep{transformer_1, transformer_2, transformer_3}. By comparing with these baseline models, we confirm that our method can effectively address the problems above and achieves better results.

In summary, the contributions of this paper are:

\begin{itemize}
  \item We fully utilize the detailed pixel-level information in RS data to improve prediction accuracy.
  
  \item We use an attention mechanism to effectively solve the ``mixed pixel'' problem.
  
  \item Our model achieves promising results on U.S. Corn Belt data, demonstrating its superiority over traditional machine learning and other deep learning models.
\end{itemize}

This paper is organized as follows. \autoref{Introduction} introduces the background on corn yield prediction, the challenge, and the objective of this paper. \autoref{Data_acquisition} presents the study area, input data, experiment design, and the evaluation metrics. \autoref{Methodology} describes the overview of the proposed attention-weighted MIL method including problem definition, attention mechanism, and MIL sampling method. \autoref{Experiment} shows the results of the experiment, discusses some abnormal results in extreme weather, and gives some detailed analysis from both spatial and temporal perspectives. \autoref{Discussion} delves into attention analysis and correlation of attention and agricultural features. Finally, \autoref{Conclusion} draws conclusions of this research.

\section{Data acquisition}
\label{Data_acquisition}

In our experiment, features are drawn from a range of sources (\autoref{tab:data}), including satellite imagery, weather, and soil properties, \etc We also incorporate the prediction year as a feature to enhance the model's understanding of temporal patterns, and use the historical average yield to serve as a baseline to improve the model's robustness.

\clearpage
\begin{table*}[ht]
\centering
\scalebox{0.55}{
\begin{tabular}{|c|c|c|c|c|c|c|}
\hline
\textbf{Category} & \textbf{Variables} & \textbf{Unit} & \makecell[c]{\textbf{Related}\\\textbf{Properties}} &  \makecell[c]{\textbf{Spatial}\\\textbf{Resolution}} & \textbf{Source} & \textbf{Latency} \\
\hline
\multirow{5}{*}{{\shortstack{\\Satellite\\Imagery}}} & Enhanced Vegetation Index (EVI) & \multirow{3}{*}{N/A} & \multirow{3}{*}{Plant vigor} & \multirow{3}{*}{500$\mathrm{m}$} & \multirow{5}{*}{MODIS} & \multirow{11}{*}{One day}\\
&Green Chlorophyll Index (GCI) &  &  &  &  &  \\
&Normalized Difference Water Index (NDWI) &  &  &  &  &  \\
 \cline{3-5} 
&Daytime Land Surface Temperature (LSTday)   & \multirow{2}{*}{Kelvin} & & \multirow{2}{*}{1$\mathrm{km}$}  &  &  \\
&Nighttime Land Surface Temperature (LSTnight) &  &  &  &  &  \\
\cline{1-3}\cline{5-5}
\multirow{6}{*}{Climate} & Mean Temperature (Tmean) & \multirow{3}{*}{$^\circ C$} & Heat stress & \multirow{6}{*}{4$\mathrm{km}$} & \multirow{6}{*}{PRISM} & \\
 & Max Temperature (Tmax) &  & &  &  &  \\
 & Minimum Temperature (Tmin) &  &  &  &  &  \\
 \cline{3-4}
 & Total Precipitation (PPT) & $\mathrm{mm}$  & \multirow{3}{*}{Water stress} &  &  &  \\
\cline{3-3}
 & maximum vapor pressure deficit (VPDmax) & \multirow{2}{*}{$\mathrm{hPa}$} &  &  &  &  \\
 & minimum vapor pressure deficit (VPDmin) &  & &  &  &  \\
\hline
\multirow{3}{*}{Soil} & Available Water Holding Capacity (AWC) & $\mathrm{cm}$ & Soil water uptake & \multirow{3}{*}{30$\mathrm{m}$} & \multirow{3}{*}{SSURGO} & \multirow{4}{*}{N/A} \\
 \cline{3-4}
 & Soil Organic Matter (SOM) & $\mathrm{kg}$/$\mathrm{m}^2$ & \multirow{2}{*}{Soil nutrient uptake} &   &  &  \\
  \cline{3-3}
 & Cation Exchange Capacity (CEC) & $\mathrm{cmol}$/$\mathrm{kg}$ & &   &  &  \\
\cline{1-5}\cline{6-6}
\multirow{2}{*}{Others} & Historical average yield & t/ha & \multirow{2}{*}{N/A} & County-level & \multirow{2}{*}{USDA NASS} &  \\ \cline{3-3}\cline{5-5}
 & Prediction year & N/A &  & N/A &  &  \\ \cline{2-7}
 \hline
\end{tabular}}
\caption{This table summarizes all the data information used in our experiments.}
\label{tab:data}
\end{table*}

% \clearpage

\subsection{Study area}
\begin{figure*}[htbp]
\centering  %图片全局居中
\includegraphics[width=1\textwidth]{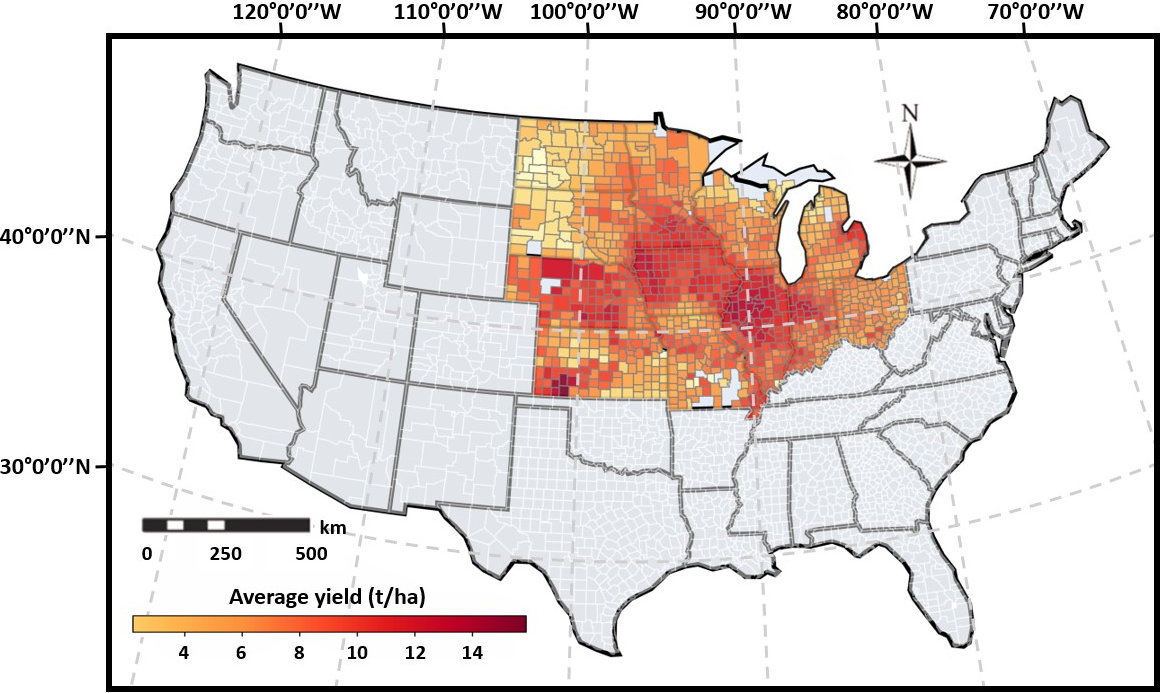}

\caption{The 5-year average yield map of each county.}
    \label{fulfig1}
\label{fig:studyarea}
\end{figure*}

The study area covers twelve states in the U.S. Corn Belt: North Dakota, South Dakota, Minnesota, Wisconsin, Iowa, Illinois, Indiana, Ohio, Missouri, Kansas, Nebraska, and Michigan. We calculate the average corn yield for all states from 2008 to 2022, as shown in \autoref{fig:studyarea}. Darker colors represent higher yields. The corn yield in the selected states accounts for most of the US total corn yield \citep{yielddata}, which is sufficient to support our predictions of corn yield for the United States. This experimental setting follows several previous works \citep{yuchi4,yuchi5,yuchi6}.

\clearpage

% \clearpage

\subsection{Satellite data}
The Moderate Resolution Imaging Spectroradiometer (MODIS) dataset \citep{modismcd} is a crucial resource for global monitoring of the earth's surface and atmosphere, captured by instruments aboard NASA's Terra and Aqua satellites. Providing data since the early 2000s, MODIS offers comprehensive coverage of Earth's landscapes, oceans, and atmosphere in 36 spectral bands. Three VIs, including Green Chlorophyll Index (GCI) \citep{gci}, Enhanced Vegetation Index (EVI) \citep{evi}, and Normalized Difference Water Index (NDWI) \citep{ndwi}, are extracted from the MODIS MCD43A4 product with 500$\mathrm{m}$ spatial resolution. GCI is a metric that assesses the chlorophyll concentration in leaves, providing insight into a plant's photosynthetic capacity. Higher values often correlate with healthier and more productive vegetation. GCI is defined as:

\begin{equation}
    GCI = \frac{NIR}{Green} - 1
\end{equation}

Where $NIR$ represents the surface reflectance in the Near Infrared band, and $Green$ represents the surface reflectance in the Green band.

EVI is an enhanced vegetation index, which optimizes the vegetation information by minimizing soil and atmosphere influences. It is particularly effective in areas with dense vegetation and provides more details about biomass and canopy structural variations:

\begin{equation}
    EVI = 2.5 \times \frac{NIR - Red}{NIR + 6 \times Red - 7.5 \times Blue + 1} 
\end{equation}

Where $Red$ and $Blue$ represent the surface reflectances of the Red band and Blue band respectively.

NDWI, the normalized difference water index, is used to monitor changes in the water content of leaves, helping to identify vegetation stress due to drought or disease. This index is particularly useful in monitoring crop conditions where precise water management is crucial:

\begin{equation}
    NDWI = \frac{NIR - SWIR}{NIR + SWIR}
\end{equation}

Where $SWIR$ represents the short-wave infrared band.

\subsection{Weather data}
The Parameter elevation Regressions on Independent Slopes Model (PRISM) dataset \citep{prism1,prism2} is a highly sophisticated climate data product that offers detailed information on various climatic elements, including precipitation, temperature, and humidity, across the United States. The weather observations used for the experiment include daily mean air temperature (Tmean), maximum air temperature (Tmax), minimum air temperature (Tmin), maximum vapor pressure deficit (VPDmax), minimum vapor pressure deficit (VPDmin), and total precipitation (PPT) from the PRISM dataset with 4 $\mathrm{km}$ spatial resolution. Tmean represents the average air temperature over a 24-hour period. Tmax and Tmin represent the highest and lowest air temperatures recorded in a day, respectively. VPDmax and VPDmin reflect the maximum and minimum differences respectively, between the amount of moisture in the air and the amount it can hold when it is saturated. PPT indicates the total amount of rain, snow, or other precipitation falling in a given time period. Moreover, daytime and nighttime land surface Temperature (LSTday and LSTnight) are extracted from the MODIS MYD11A2 product \citep{modismyd} with 1 $\mathrm{km}$ spatial resolution. LSTday and LSTnight represent the temperature of the Earth's surface during the daylight and night hours. These meteorological features are crucial in understanding weather patterns and can have significant impacts on agricultural and environmental conditions.

\subsection{Soil data}
For soil properties, we use Available Water Holding Capacity (AWC), Soil Organic Matter (SOM) and Cation Exchange Capacity (CEC) from Soil Survey Geographic database (SSURGO) \citep{ssurgo} with 30$\mathrm{m}$ spatial resolution. AWC represents the soil's ability to hold water available for plant use, SOM is a measure of the amount of organic material present in the soil, and CEC indicates the soil's ability to hold cations, thereby influencing soil fertility and nutrient availability. 

\subsection{Cornfield mask}
A CDL (Cropland Data Layer) \citep{cdl} mask is a digital raster-based dataset primarily used for identifying and categorizing crop-specific land cover over agricultural regions. It is derived from the Cropland Data Layer, an annual publication by the United States Department of Agriculture's National Agricultural Statistics Service (USDA-NASS), which utilizes satellite imagery to produce high-resolution, geo-referenced maps detailing the types of crops planted in specific areas. We use the CDL mask to extract the cornfield lands with 30$\mathrm{m}$ spatial resolution.

\subsection{Data preprocessing}
On the Google Earth Engine (GEE) platform \citep{gee}, we spatially aggregate all the RS and weather features (GCI, EVI, NDWI, LSTday, LSTnight, Tmean, Tmax, Tmin, PPT, VPDmin, VPDmax, AWC, SOM, CEC). These features are then grouped temporally into 16-day intervals from mid-May to early October to cover the corn-growing period. The dataset covers the years 2008 to 2022 and includes county-level historical yield records obtained from the USDA NASS \citep{yielddata}. Furthermore, our model integrates the year data to capture specific time-related features and calculates the 5-year historical average yield to provide a baseline for model robustness. The feature length of each input data point is 159. After processing, the dataset includes approximately 600 to 800 counties per year, with each county represented by 100 pixels. The test set is from 2018 to 2022, while the training set ranges from 2008 to the year preceding the testing year. We set aside 20\% of the training data for model validation.

% \clearpage
\section{Methodology}
\label{Methodology}

\begin{figure*}[htbp]
\centering  %图片全局居中
\includegraphics[width=1\textwidth]{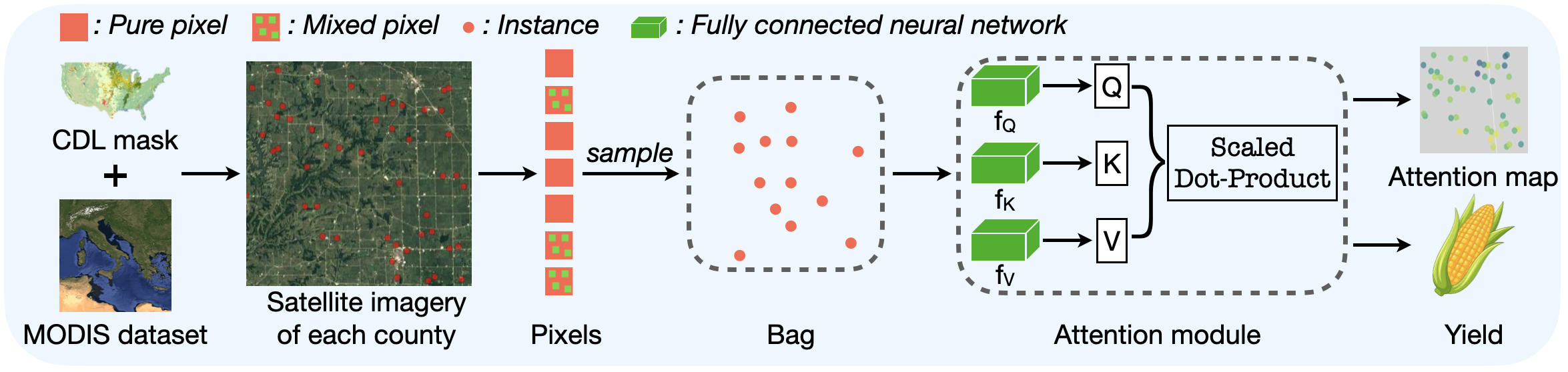}
\caption{The pipeline of the proposed method.}
\label{fig:pipeline}
\end{figure*}

In \autoref{fig:pipeline}, our method is structured as follows: we begin by applying the CDL mask to the MODIS dataset. This is followed by selecting some pixels within a county for use in training the model and making predictions. Next, we employ MIL to establish a many-to-one mapping. Finally, we implement an attention mechanism to assign weights to these pixels, expecting to assign higher weights to pure pixels and lower weights to mixed pixels.

\clearpage

% \clearpage

In this section, we will 1) formally define the county-level corn yield prediction (\autoref{Problem definition}); 2) describe the MIL method and how to apply it to county-level corn yield prediction with individual pixels' information input (\autoref{Multiple instance learning}); 3) introduce the attention mechanism in detail and explain its application in our setting (\autoref{Attention mechanism implementation}).

\subsection{Problem definition}
\label{Problem definition}
The county-level corn yield prediction process involves several key steps. First, the USDA NASS Cropland Data Layer (CDL) mask \citep{cdl} is used to identify and extract the cornfields at 30$\mathrm{m}$ resolution. Subsequently, feature datasets (satellite imagery, climate, soil, \etc) will be processed to the same format. Finally, we use these county-level features as an input and county yield as output to train a machine learning model.

To formalize the county level crop yield prediction problem,  let $n$ be the number of counties in the given study area. Let $I_i$ be the satellite imagery and  $y_i$ be the yield for each county $i$, $i = 1,...,n$. Each image is on the space $\mathbb{R}^{W \times H \times C}$, where $W$ and $H$ are the width and height of the image, and $C$ denotes the channels. We calculate VIs from the satellite imagery, along with other agricultural and environmental features (weather, soil properties, \etc), to form our county-level input feature $F_i$. From a machine learning perspective, our training objective is to develop a model $f$, $f(F_i) = y_i$.

\subsection{Multiple instance learning}
\label{Multiple instance learning} 

Conventional county-level corn yield prediction methods usually aggregate all the pixels in one county into a single value \citep{yuchi1, yuchi7, yuchi8}, which may compromise the integrity of county data representation, potentially losing some detailed information. In response to these limitations, we investigate pixel-level feature processing and apply MIL to the task. In real-world applications, there are often scenarios where a single label represents a group of data points. In MIL, this setting is described as a \emph{bag} containing many \emph{instances}, from which we need to predict the label. Thus, unlike the traditional one-to-one mapping in machine learning, MIL involves a many-to-one mapping. Additionally, MIL usually assumes that there is some connection between instances within a bag or that the bag has some kind of special structure. \eg, to fully utilize the connection of instances in a bag, Cluster-MIL \citep{clustermir} assumes that individual instances come from a group of underlying clusters and employs the MIL model to learn the internal structure of bags that contain instances from various unknown distributions. 

In our work, we follow the approach of \citep{attmir}, using an attention module to process each county and learning the connections between pixels within the county to improve the model performance. One key difference is that their work focuses on classification, whereas ours focuses on regression. We will first provide a detailed description of how attention is used to perform the MIL classification task. 

\textbf{Problem formulation for MIL in classification task} 

In traditional binary supervised learning, the goal is to find a model that predicts the value of a target variable, $y \in \{0, 1\}$, for a given instance, $\mathbf{x} \in \mathbb{R}^D$. However, in the MIL problem there is a bag of instances, $X = \{\mathbf{x}_1, \ldots, \mathbf{x}_K\}$. Each bag has a single binary label $Y$ associated with it. Additionally, It is assumed that individual labels exist for the instances within a bag, i.e., $y_1, \ldots, y_K$ and $y_k \in \{0, 1\}$, for $k = 1, \ldots, K$, but these labels are inaccessible and remain unknown during training.

The assumptions of the MIL problem can be expressed as follows:

\begin{equation}
Y = \begin{cases} 
0, & \text{if} \ \sum_k y_k = 0, \\
1, & \text{otherwise}.
\end{cases}
\end{equation}

or it can be re-formulated as:

\begin{equation}
Y = \max_k \{y_k\}.
\end{equation}

These assumptions imply that an MIL model must be permutation-invariant \citep{zaheer2017deep, qi2017pointnet}. The key component in constructing an MIL model is to combine the instances with a permutation-invariant function $\sigma$, which is also referred to as MIL pooling. 

\citet{attmir} proposed using an attention module \citep{attention} to weight the instances using neural network. We will explain the attention mechanism in detail later (\autoref{Attention mechanism implementation}). The weighted average function meets the requirements for being permutation-invariant. Given a bag \(X\) of \( K \) instances, \( X = \{\mathbf{x}_1, \ldots, \mathbf{x}_K\} \), the following MIL pooling ($\sigma$) is proposed:

\begin{equation}
\mathbf{z} = \sum_{k=1}^K a_k \mathbf{x}_k,
\end{equation}

where

\begin{equation}
a_k = \frac{\exp\left(\mathbf{w}^\top \tanh\left(\mathbf{V} \mathbf{x}_k^\top\right)\right)}{\sum_{j=1}^K \exp\left(\mathbf{w}^\top \tanh\left(\mathbf{V} \mathbf{x}_j^\top\right)\right)},
\end{equation}

where \( \mathbf{w} \in \mathbb{R}^{L \times 1} \) and \( \mathbf{V} \in \mathbb{R}^{L \times M} \) are parameters. Moreover, the hyperbolic tangent \( \tanh(\cdot) \) element-wise non-linearity is used for both positive and negative values, ensuring proper gradient flow. This construction enables the discovery of connections among instances. Afterwards, when the permutation-invariant function \(z\) has been constructed , any classification module can be applied to predict the class of the bag. Ideally, when the label \( Y = 1 \), high attention weights should be assigned to instances likely to have the label \( y_k = 1 \). 

\textbf{Problem formulation for MIL in regression task} For the county-level corn yield prediction task, we have the yield (label) of each county (bag), and each county is composed of a number of pixels (instances). By utilizing the detailed information of each county, MIL can improve prediction accuracy. Suppose there are $m_i$ pixels in each county $i$, $i=1,...,n$. To differentiate from the county-level feature $F$, we use $P$ to represent pixel-level features. For each pixel $j$, $j=1,...,m_i$, we extract its feature $P^j_i$. There exists a model $f$ with many-to-one mapping, namely $f(P^{1}_i, P^{2}_i,... \ldots , P^{m_i}_i) = y_i$. We hope our model can automatically learn the structure of a county as well as the connections between different pixels within it, thereby improving corn yield predictions.

A key step in MIL is selecting representative instances within each bag, since training on all instances in a bag is computationally expensive, especially for large bags. Additionally, without clear indications of which instances are more influential for a bag's label, treating all instances with equal importance could mislead the model during the learning process. In our county-level corn yield prediction task, we utilize a unique sampling method, drawing inspiration from the mixed pixel problem, to select the most representative pixels within each county. As illustrated in \autoref{fig:mixedmodis}, the VIs of a CDL pixel are derived from the information of the encompassing MODIS pixel; all the CDL pixels within one MODIS pixel share the same VI. We believe that within a large MODIS pixel, we only need to select one CDL pixel to represent it; the rest are redundant. Therefore, in the MIL sampling process, we randomly select one CDL pixel from each MODIS pixel, removing the remaining pixels with duplicate VIs. 

To simplify the learning process, we need to sample an appropriate number of pixels in each county. On the one hand, selecting fewer pixels might not capture the complete information in each county. On the other hand, over-sampling may result in an insufficient number of qualified counties, making it impossible to obtain sufficient data for training. Here, we randomly select 100 pixels from each county $i$, $i=1,...,n$, denoted as $\hat{P_i} = (P^1_i,P^2_i,...,P^{100}_i)$. Our experiments in Section 4.4 have shown that these selected pixels will adequately capture the county's information, as will be demonstrated in the experiment (\autoref{The number of instances in a bag}). For simplicity of notation, we omit the subscript $i$ of $\hat{P_i}$ here, so it becomes $\hat{P}$. We aim to train a model $f$, $f(\hat{P})=y_i$, that can predict the yield accurately.

\subsection{Attention mechanism implementation}
\label{Attention mechanism implementation}

In previous sections, we discuss the mixed pixel problem, where individual CDL pixels contain a mixture of different land types information. This issue is particularly relevant when dealing with datasets of varying spatial resolutions, making it difficult to manually integrate and weight these mixed pixels effectively. To overcome these challenges, we employ an attention mechanism. The attention mechanism automatically assigns higher weights to pixels whose information is predominantly composed of cornfield and lower weights to those with more irrelevant land types.

The attention mechanism \citep{attention} in machine learning assigns varying importance to different input elements, enabling the model to focus on inputs that are more relevant to the output \citep{attentionsurvey}. This helps models perform better in complex tasks by learning context-specific representations. In our task, we not only use attention to implement MIL, but also to assign appropriate weights to the pixels in each county, hoping that attention can automatically determine the significance of each pixel. Ideally, it assigns higher weights for pure pixels and lower weights for mixed pixels. Suppose there are $100$ pixels in county $i$, $i=1,...,n$. For the feature $P_i^j$ of pixel $j$, $j=1,...,100$, our goal is to learn a weight $W_i^j$. We are aiming for such a model $f$, $f({W^1_i}^\top P^{1}_i, {W^2_i}^\top P^{2}_i,..., {W^{100}_i}^\top P^{100}_i) = y_i$. In this way, we can reduce as much the noisy information in mixed pixels as possible. 

The neural network learns patterns from input data through a process called forward propagation by some layers and performs error adjustment with a backpropagation process \citep{dlbook}. In the attention module, we use $100$ pixels from each county $i$ and extract their features $P^j_i$ as the input vector $\hat{P_i} = (P^1_i,P^2_i,...,P^{100}_i)$.  For simplicity of notation, we omit the subscript $i$ of $\hat{P_i}$ here, so it becomes $\hat{P}$. Then it is passed through three separate modules $f_Q$, $f_K$, and $f_V$ to obtain the respective vectors: query $Q$, key $K$, and value $V$. The three separate modules are fully connected layers, which are a fundamental building block in artificial neural networks. In these layers, each neuron is connected to every neuron in the previous layer. Each connection has an associated weight, which determines the strength and direction (positive or negative) of the influence of the input on the output. Additionally, each output neuron has a bias term that is added to the weighted sum of inputs before applying the activation function. The output of a fully connected layer is computed by taking the dot product of the input vector and the weight matrix, adding the bias vector, and then applying an activation function. 

\begin{align}
    Q = f_Q(\hat{P}), K = f_K(\hat{P}), V = f_V(\hat{P})
 \end{align}

Then we calculate the attention score on a batch of queries $Q$, keys $K$, and values $V$ concurrently, the output matrix is computed as follows:

\begin{align}
    Attention(Q, K, V) = softmax(\frac{QK^\top}{\sqrt{d_K}})V
\end{align}

Where $d_K$ is the dimension of $K$. 

And The softmax function \citep{dlbook} is defined as:

\begin{align}
\text{softmax}(\mathbf{z})_i = \frac{e^{z_i}}{\sum_{j=1}^{K} e^{z_j}}
\end{align}

Where \(\mathbf{z} = (z_1, \dots, z_K)\) is the input vector for \( i = 1, \dots, K \). \( e^{z_i} \) is the exponential of the \( i \)-th element of the input vector \(\mathbf{z}\). 

The similarity between the query $Q$ and the key $K$ is captured through their dot product. The higher the dot product, the more similar the query $Q$ is to the key $K$. The dot product result is then passed through a softmax function to obtain the weights, which range between 0 and 1 and sum to 1. The softmax function ensures that the larger dot products get higher weights. Finally, these softmax weights are used to compute a weighted sum of the values $V$. Each value is multiplied by its corresponding weight, and these products are then summed. This final result is the output of the attention mechanism, which represents the context relevant to the query $Q$ given key-value pairs.

By combining the above equations, we denote the function of our attention module as $Att(\hat{P})$. Then our model $f$ becomes $f(Att(\hat{P}))=y_i$.

Using attention to weight the pixels in each county allows us to achieve MIL, enabling the learning of each county’s detailed information. This method also helps in determining the completeness of information from different pixels, effectively filtering out noisy data in mixed pixels.

% \clearpage
\section{Experiment}
\label{Experiment}

\subsection{Experimental setup}
In order to evaluate the performance of the proposed model, three traditional machine learning models, including linear regression (LR) \citep{prml}, ridge regression (RR) \citep{ridge}, and random forest (RF) \citep{rf}, are selected as baseline models for comparison. In particular, LR is a statistical method for predicting an outcome based on the linear relationship with one or more independent variables. RR is a variant of LR that adds a penalty term to prevent overfitting in high-dimensional datasets. RF is an ensemble machine learning algorithm that aggregates many decision trees to achieve more accurate predictions. These models are implemented using scikit-learn \citep{sklearn} with default parameters. To highlight the role of MIL, Ins-MIL \citep{insmir} is also included as a comparison, which employs each instance for regression directly, without taking the bag structure into account. Additionally, based on our review of the application of deep learning models in remote sensing (\autoref{Introduction}), we select the current mainstream models—CNN \citep{cnn}, LSTM \citep{lstm}, ResNet \cite{resnet}, and Transformer \citep{attention}. Specifically, for the CNN model, since we are not directly inputting 2D remote sensing images but rather a temporal vector, we replace the traditional 2D convolution with 1D convolution. For LSTM, ResNet, and Transformer, we use PyTorch's built-in model \citep{pytorch} and adjust the input and output to fit our data format. To use the PyTorch built-in ResNet18 model with 1D data, we adapt the 1D input to match the expected 2D input format, which is originally designed for images. This involves reshaping the input data, modifying the first convolutional layer, and replacing it to accept a single input channel. For all the deep learning models we use Adam \citep{adam} as the optimizer with a batch size of $100$. Initially, the learning rate is set to $0.1$ and is reduced by a factor of 10 whenever the error reaches a plateau. All experimental results are obtained after averaging five repetitions. For each experiment, we randomly select different random seeds to ensure the results are fair and reproducible.

\subsubsection{Performance Metrics}
The experiment results are measured by standard error metrics $RMSE$ and $R^2$ \citep{rmse}:

\begin{equation}
    RMSE = \sqrt{\frac{\sum_{i=1}^n(y_i-\hat{y}_i)^2}{n}}
\end{equation}

\begin{equation}
    R^2 = 1 - \frac{\sum_{i=1}^n(y_i-\hat{y}_i)_2}{\sum_{i=1}^n(y_i-\bar{y})^2}
\end{equation}

where n is the total number of observations (counties or pixels). $y_i$ is the predicted value (predicted yield) for the $i\textsuperscript{th}$ observation (one pixel in RS imagery).
$\hat{y}_i$ is the actual value (observed yield) for the $i\textsuperscript{th}$ observation. $\bar{y}$ is the mean of the observed data. The observed yield data are the official historical reports of the county level yields from USDA, NASS \citep{yielddata}.

\subsection{Model Performance Evaluation}
The proposed model, MIL with attention mechanism (Att-MIL), is compared with other methods in \autoref{tab:result}. The results show that our approach outperforms all other methods across all years and metrics. The superior performance of our approach can be attributed to its effective extraction and incorporation of inter-pixel correlations within each county, which other methods overlook. The second-best method, Ins-MIL, effectively utilizes the MIL concept and recognizes that each county consists of multiple pixels. This understanding provides a more comprehensive view than merely considering each county as a single entity and results in better performance compared to traditional regression-based methods. For the traditional machine learning models, RF, ranking third, exhibits relatively good performance due to its robustness to noise and ability to model non-linear relationships. However, it does not inherently model detailed county information, thereby performing worse than the MIL-based methods. LR and RR are based on the assumptions of a linear relationship, which are not entirely accurate in the context of county-level corn yield prediction. These two methods exhibit the weakest performance in our experiments. 

For the deep learning-based models, ResNet18 performs the worst. This is likely due to the significant difference between the pre-trained dataset and our dataset, as well as the fact that ResNet is not designed for 1D features. The CNN model performs slightly better. Although we modify the 2D convolution to 1D to adapt to our input data, the focus of CNN on local correlations limits its effectiveness on our dataset. The Transformer model comes next, as its large architecture requires substantial data to avoid overfitting, and our data volume is clearly insufficient. The LSTM model performs the best and is closest to the performance of our method. LSTM model excels in capturing long-term temporal dependencies and sequential patterns, which are crucial for yield prediction, as seasonal and temporal variations significantly impact results. The above comparisons underscore the benefits of integrating the attention mechanism and MIL in tackling the complexity of county-level corn yield prediction task.

\clearpage

\begin{table}[ht]
\centering
\scalebox{0.7}{
\begin{tabular}{ccccccccccc}
\toprule
\diagbox{Year}{Method} & \multicolumn{2}{c}{Att-MIL} & \multicolumn{2}{c}{Ins-MIL} & \multicolumn{2}{c}{LR} & \multicolumn{2}{c}{Ridge} & \multicolumn{2}{c}{RF} \\
\midrule
 & $RMSE$ & $R^2$& $RMSE$ & $R^2$& $RMSE$ & $R^2$& $RMSE$ & $R^2$& $RMSE$ & $R^2$\\
\hline
2018 & \textbf{1.00} & \textbf{0.67} & 1.42 & 0.50 & 1.49 & 0.45 & 1.49 & 0.45 & 1.41 & 0.50\\
2019 & \textbf{0.99} & \textbf{0.58} & 1.18 & 0.51 & 1.46 & 0.26 & 1.45 & 0.27 & 1.60 & 0.10\\
2020 & \textbf{1.00} & \textbf{0.48} & 1.35 & 0.20 & 1.53 & -0.02 & 1.81 & -0.42 & 1.41 & 0.13\\
2021 & \textbf{0.99} & \textbf{0.64} & 1.16 & 0.60 & 1.78 & 0.05 & 2.04 & -0.24 & 1.08 & 0.64\\
2022 & \textbf{0.83} & \textbf{0.84} & 1.07 & 0.73 & 1.35 & 0.65 & 1.18 & 0.73  & 1.19 & 0.73\\
\bottomrule
\end{tabular}}

\bigskip

\scalebox{0.7}{
\begin{tabular}{ccccccccc}
\toprule
\diagbox{Year}{Method} & \multicolumn{2}{c}{CNN} & \multicolumn{2}{c}{LSTM} & \multicolumn{2}{c}{Resnet18} & \multicolumn{2}{c}{Transformer}  \\
\midrule
 & $RMSE$ & $R^2$& $RMSE$ & $R^2$& $RMSE$ & $R^2$& $RMSE$ & $R^2$\\
\hline
2018 & 1.48 & 0.42 & 1.03 & 0.62 & 1.67 & 0.42 & 1.12 & 0.55  \\
2019 & 1.34 & 0.44 & 1.01 & 0.52 & 1.53 & 0.32 & 1.15 & 0.47 \\
2020 & 1.33 & 0.21 & 1.02 & 0.42 & 1.73 & 0.27 & 1.07 & 0.41 \\
2021 & 1.41 & 0.47 & 1.06 & 0.57 & 1.67 & 0.35 & 1.14 & 0.51  \\
2022 & 1.42 & 0.62 & 0.91 & 0.79 & 1.32 & 0.45 & 0.97 & 0.77  \\
\bottomrule
\end{tabular}}

\caption{Model evaluation results in 2018-2022.}
\label{tab:result}
\end{table}

% \clearpage

\begin{figure}
    \captionsetup[subfigure]{labelformat=empty}

    \tikzset{inner sep=0pt}
    \setkeys{Gin}{width=1\textwidth}
    \centering
\subfloat[\label{mysubfig1}]{%
\tikz{\node (a) {\includegraphics{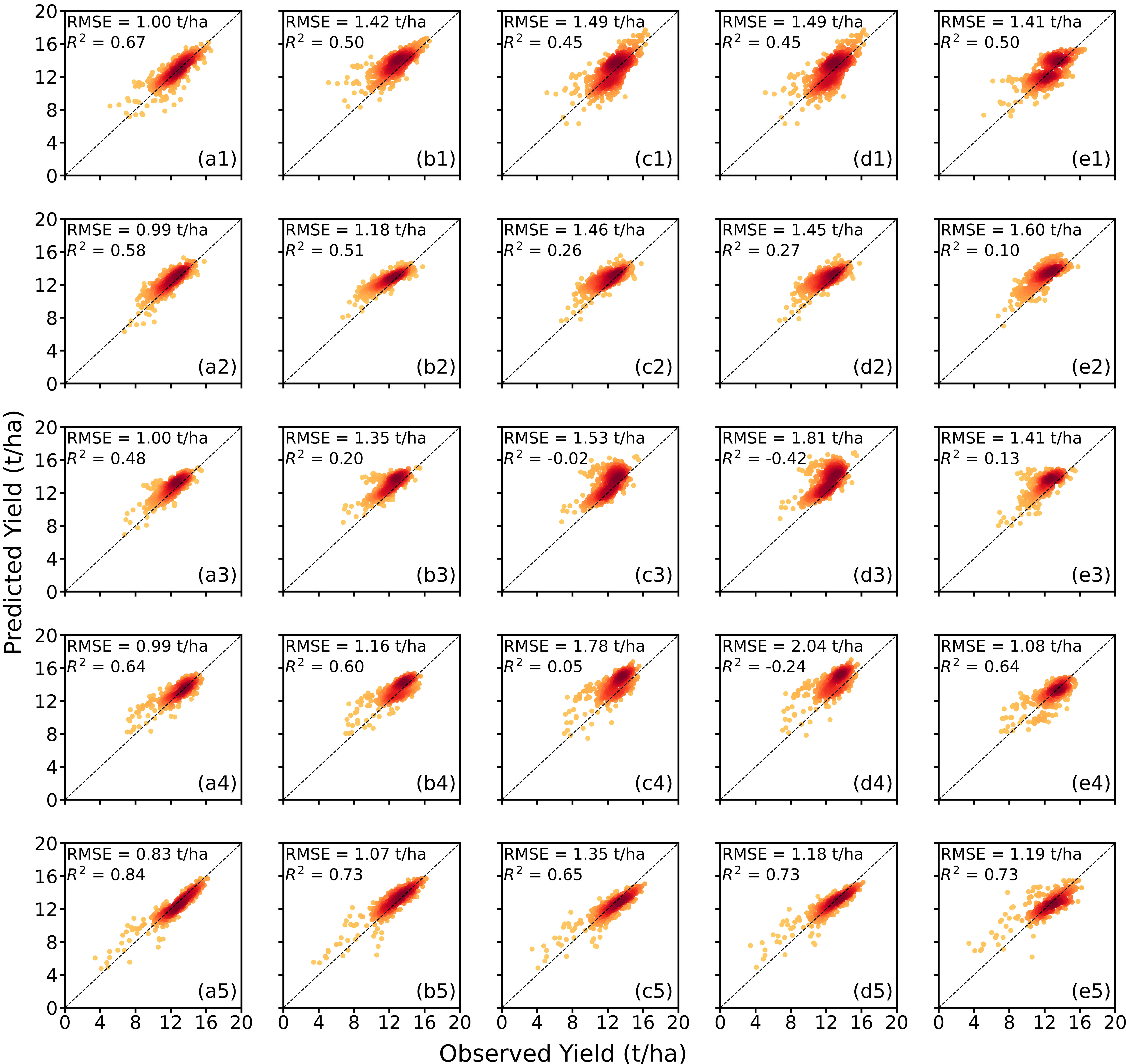}};
      }}
      
\vspace{-1em}

% \subfloat[\label{mysubfig2}]{%
% \tikz{\node (a) {\includegraphics[width=2.5in]{./fig/scatter/scatter_colorbar.png}};
%       }}

\caption{Scatter plots of observed yields vs. predicted yields of (a) Att-MIL, (b) Ins-MIL, (c) LR, (d) Ridge, (e) RF in 5 testing years: (1) 2018; (2) 2019; (3) 2020; (4) 2021; (5) 2022. The colorbar represents the density of points at each location.}
\label{fig:scatter}
\end{figure}

To illustrate the relationship between predicted and reported yields, we use scatter plots to provide a clear visual representation of their alignment (\autoref{fig:scatter}). Each point in the figures corresponds to a specific county. From these scatter plots, we observe that predictions made by Att-MIL are tightly gathered around the line of perfect fit. The scatter plot of Ins-MIL, although less gathered than that of our method, still demonstrates a significant degree of accuracy. RF does not capture detailed information within counties; hence, its results on the scatter plot are more scattered compared to the MIL-based methods. In contrast, LR and RR display the greatest spread of data points. This indicates that our approach provides more consistent and accurate results and outperforms the other methods.

\clearpage

% \clearpage

\begin{figure}
    \captionsetup[subfigure]{labelformat=empty}

    \tikzset{inner sep=0pt}
    \setkeys{Gin}{width=1\textwidth}
    \centering
\subfloat[\label{mysubfig1}]{%
\tikz{\node (a) {\includegraphics{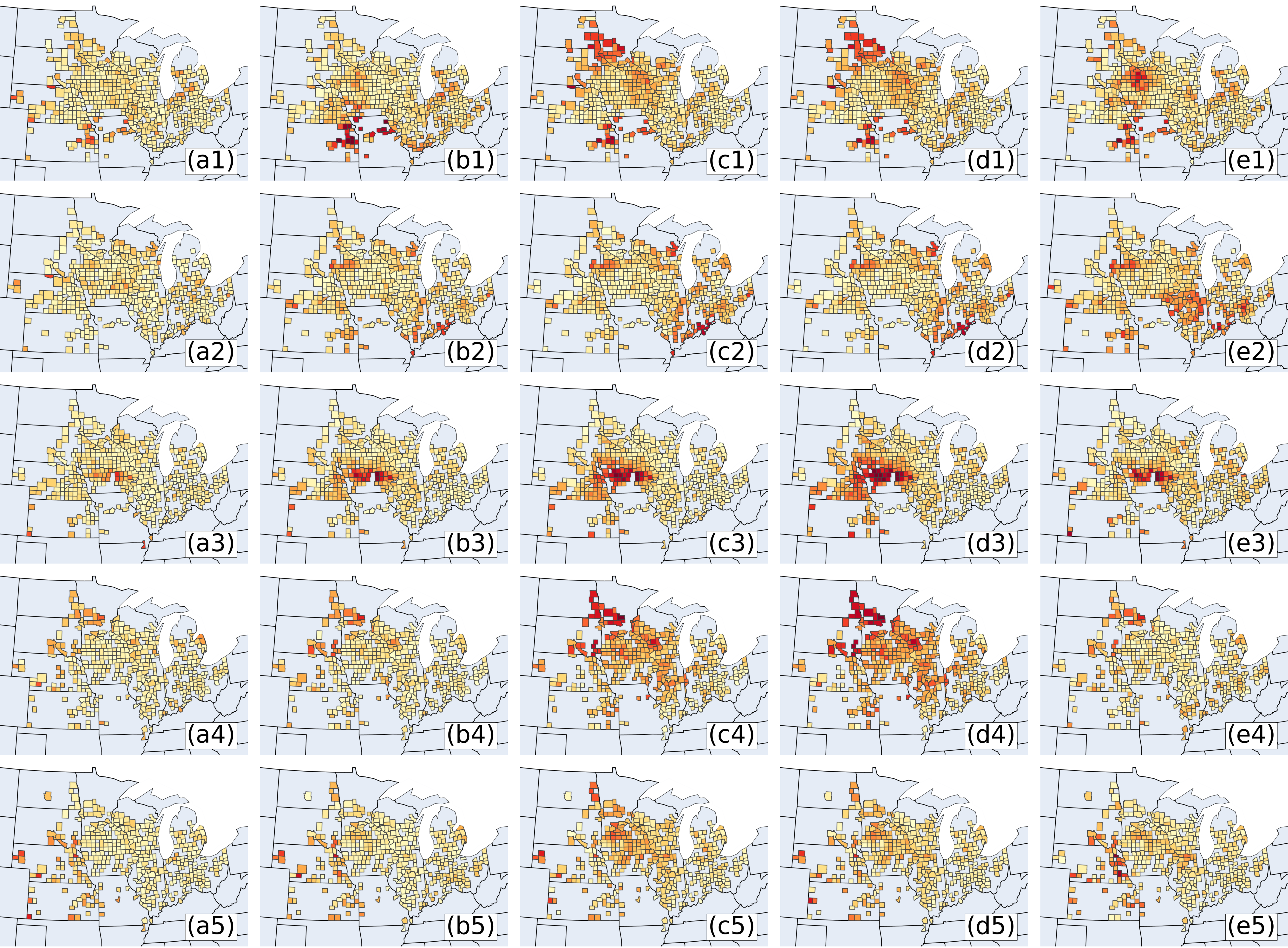}};
      }}
      
\vspace{0em}

\subfloat[\label{mysubfig2}]{%
\tikz{\node (a) {\includegraphics[width=3in]{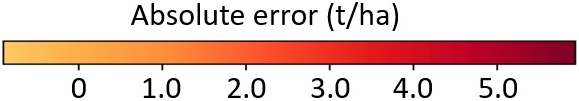}};
      }}

\caption{Absolute error maps of (a) Att-MIL, (b) Ins-MIL, (c) LR, (d) Ridge, (e) RF in 5 testing years: (1) 2018; (2) 2019; (3) 2020; (4) 2021; (5) 2022.}
\label{fig:abserr}
\end{figure}

Next, to visualize the magnitude of prediction errors spatially across different regions, we present a choropleth map that illustrates the absolute error of each method across various states (\autoref{fig:abserr}). The absolute error map provides a geographic representation of the prediction performance of each method at the county level. On the map, each colored block represents a county, with color intensity indicating the prediction error magnitude, deeper colors signify higher errors. Upon examination, it becomes clear that our method presents the lightest color across most counties. This suggests that our approach achieves the lowest absolute error overall, indicating superior prediction performance.

\clearpage

% \clearpage

\begin{figure*}[htbp]
\centering  %图片全局居中
\includegraphics[width=0.4\textwidth]{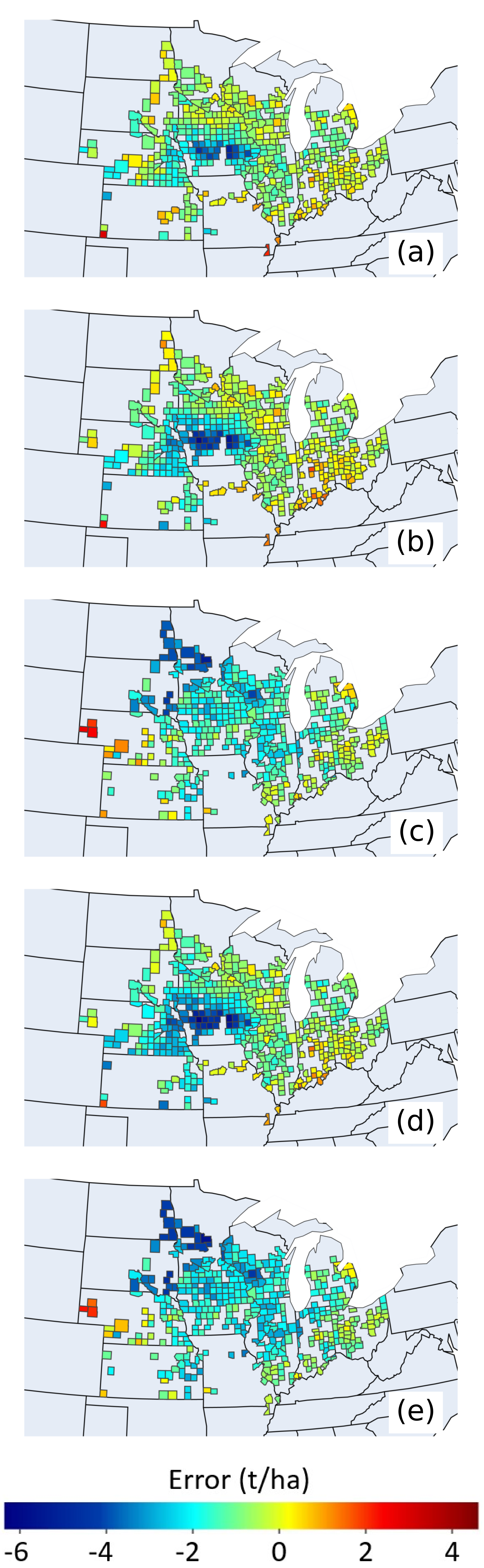}
\caption{The error maps in some abnormal years: (a) Ins-MIL in 2020, (b) LR in 2020, (c) LR in 2021, (d) Ridge in 2020, (e) Ridge in 2021. A negative value indicates overestimation, while a positive value indicates underestimation.}
\label{fig:err}
\end{figure*}

In comparison, the map corresponding to Ins-MIL exhibits slightly deeper colors, especially in Kansas in 2018 and in Iowa in 2020. This is because 2018 was particularly dry for many parts of Kansas, especially during key growth stages of corn \citep{kansas2018yield}. The U.S. Drought Monitor observed parts of Kansas in severe to extreme drought conditions, particularly in the earlier part of the year \citep{kansas2018drought}. Furthermore, on August 10, 2020, a severe windstorm known as a ``derecho'' swept through Iowa \citep{iowa2020stormnws}. This storm brought extremely high winds, with some reports of wind speeds reaching over 100$mph$ in certain areas. The derecho flattened cornfields, broke plants, and caused widespread destruction to crops across the central part of the state. This event had a significant negative impact on corn yields in Iowa for 2020. We use the error map to validate the connection between the storm and prediction error (\autoref{fig:err}). For cornfields affected by the storm, the satellite imagery does not change significantly in a short period of time. As a result, machine learning models tend to overestimate the yield of those affected areas. Indeed, in 2020, the results for this region show an overestimation, which is consistent with our hypothesis. The color intensities of these blocks in the maps of RF, LR, and RR are even deeper in Kansas and Iowa. Besides appearing darker in the two aforementioned regions, these maps also show extensive dark areas at the intersection of North Dakota, Minnesota, and South Dakota in 2021. This is because, in 2021, much of the upper Midwest, including North Dakota, South Dakota, and Minnesota, experienced severe drought conditions \citep{nd2021drought,sd2021drought,mn2021drought,mn2021drought1}. In particular, North Dakota also experienced higher-than-average temperatures during the growing season \citep{nd2021temp}.  Additionally, hot weather does not immediately affect satellite imagery, but it will impact the internal growth of plants and reduce yield. Therefore, high temperatures can also lead to corn yield overestimation, which is verified in our error map.

\clearpage

% \clearpage

\subsection{In-season yield prediction}
\begin{figure*}[htbp]
\centering  %图片全局居中
\includegraphics[width=1\textwidth]{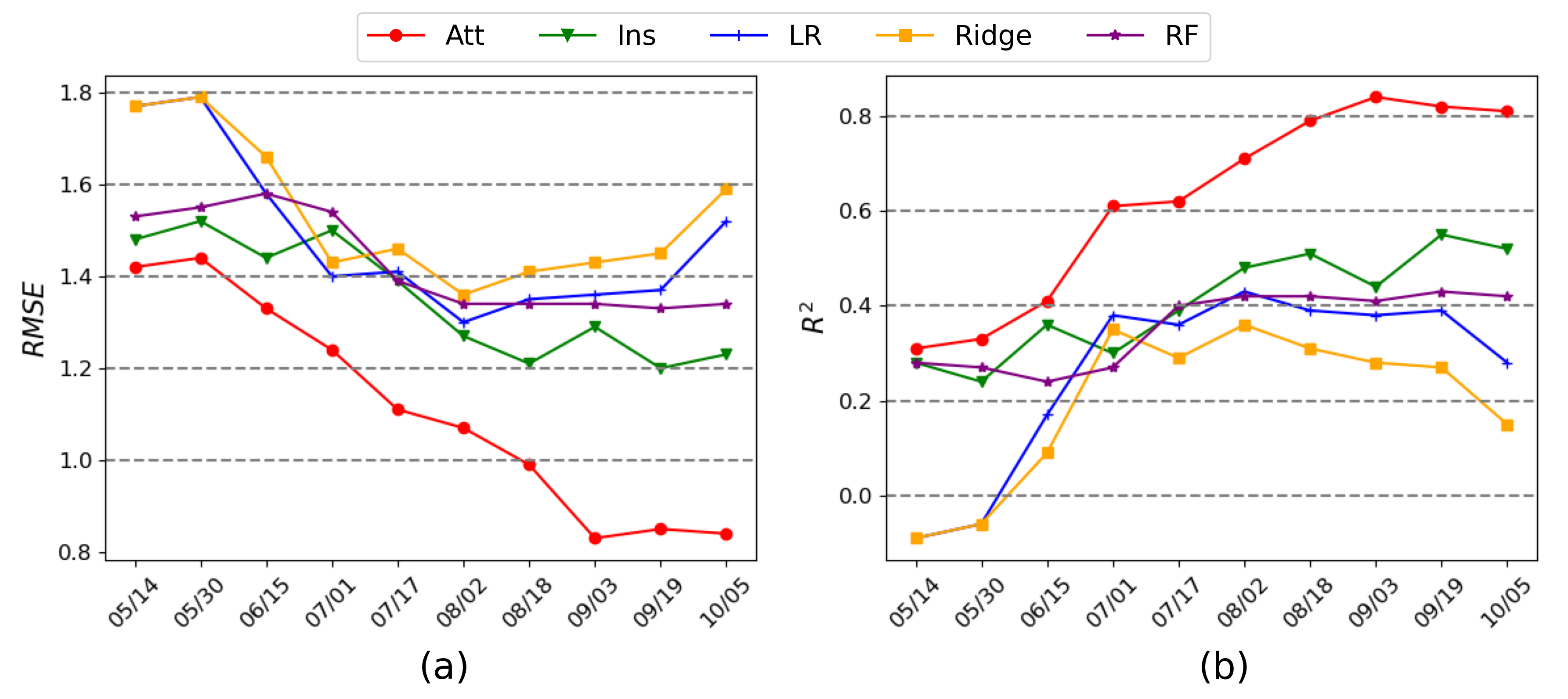}
\caption{In in-season county-level corn yield prediction in 2022: (a) $RMSE$ and (b) $R^2$.}
\label{fig:inseason}
\end{figure*}

To verify the capability of our model in providing good prediction throughout various growth stages of corn, we conduct a comparative experiment of all models for in-season county-level corn yield prediction. The timeframe is divided from mid-May to early October into 10 intervals, predicting once every 16 days (\autoref{fig:inseason}). Predicting corn yield is challenging in the initial stages of the growing season because the correlation between satellite imagery features and yield is very weak \citep{inseason}. This experiment is based on the data from 2022, with the training data spanning from 2008 to 2021. It is clear that our method incorporating attention and MIL achieves the best $RMSE$ and $R^2$ values, indicating it provides the best fit to the data across all dates. In the mid-term stage of corn growth, the performance of the four methods other than Att-MIL is roughly similar. Ins-MIL performs second-best for most of the time, while RF is somewhat inferior. LR and RR perform the worst in the early and late stages. This superior performance of our model is due to its ability to manage uncertainty and noise during the early stages of corn growth.

\clearpage

% \clearpage

\subsection{The number of instances in a bag}
\label{The number of instances in a bag}
\begin{figure*}[htbp]
\centering  %图片全局居中
\includegraphics[width=1\textwidth]{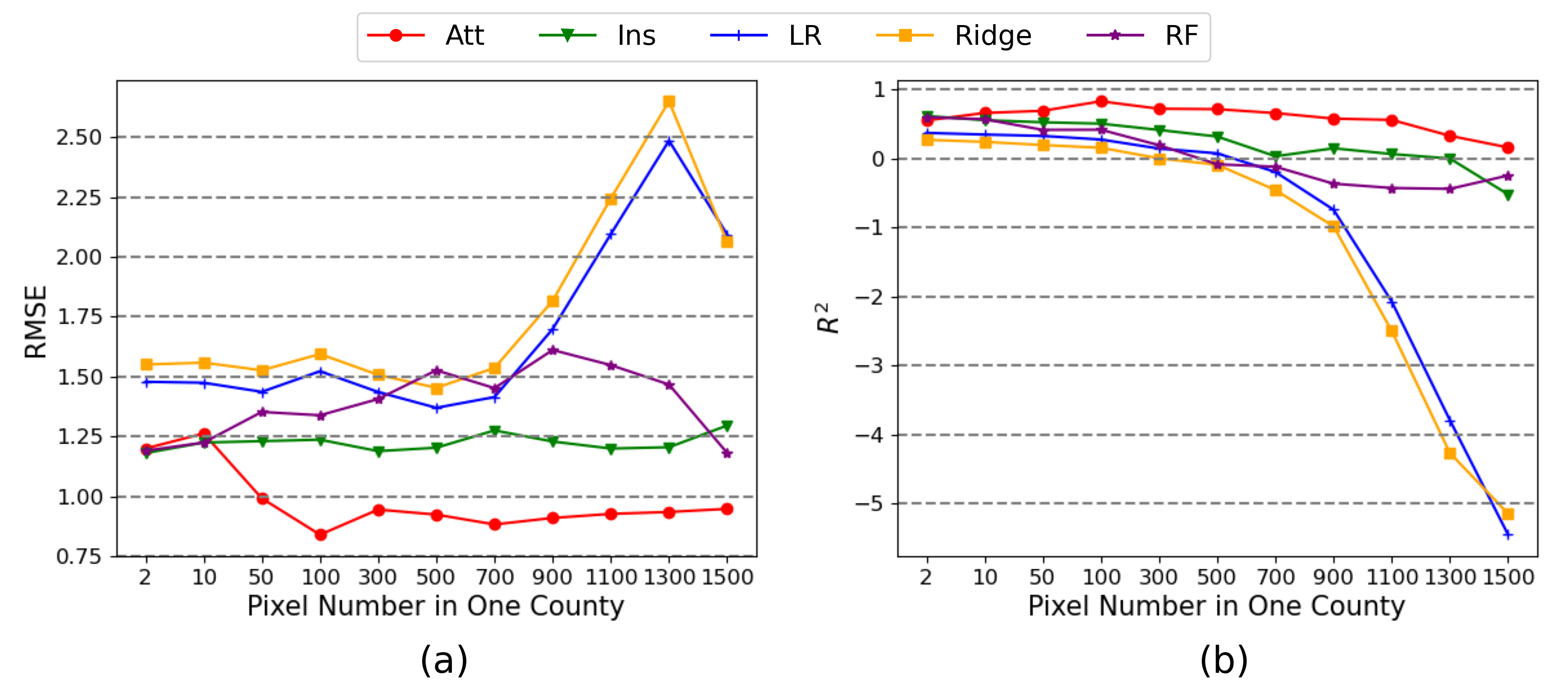}
\caption{The results of our model in 2022 under different numbers of selected pixels in one county: (a) $RMSE$ and (b) $R^2$.}
\label{fig:multibag}
\end{figure*}

In our experiment, a crucial parameter is the number of pixels selected from each county. Selecting an excessive number of pixels in each county may lead to a reduced number of qualified counties. Conversely, if the number of selected pixels is too small, they may not adequately represent the complete information of a county, negatively impacting the model's training and prediction. This naturally introduces a trade-off where more pixels in a single county lead to fewer counties being selected, and vice versa. The objective of this section is twofold: (1) identifying an optimal parameter range that allows the model to reduce the size of the dataset without sacrificing accuracy; and (2) studying the model's robustness under various parameter conditions. Therefore, we conduct a thorough experiment by varying the number of pixels per county between 2 and 1500 (\autoref{fig:multibag}). This experiment is based on the testing years from 2018 to 2022, while the training year ranges from 2008 to the year preceding the testing year.

\clearpage

% \clearpage

As shown in \autoref{fig:multibag}, our proposed Att-MIL model  consistently demonstrates superior performance, while the Ins-MIL and RF models exhibit significantly poorer performance. Meanwhile, LR and RR perform substantially worse as  the number of pixels per county increases (RMSE increases and R2 decreases). This might be attributed to these models’ limited ability to learn from larger sample datasets . Notably, even as the pixel number of a county decreases to around 100,  our model’s predictive performance remains robust. In addition, the results suggest a suitable pixel count ranges from 100 to 1000. Within this range, the predictions of our Att-MIL model exhibit minimal fluctuation, with the optimal outcome achieved around 100 pixels. These findings highlight the effectiveness and robustness of the proposed Att-MIL model, particularly in scenarios with limited data. This implies that our model is well-suited for handling datasets of varying sizes, demonstrating a significant advantage in real-world county-level corn yield prediction.

% \clearpage
\section{Discussion}
\label{Discussion}
This section provides a detailed analysis of the attention mechanism. In \autoref{corn_ratio}, we prepare the necessary data and perform data preprocessing for the analysis experiment. Next, in \autoref{attention_analysis}, we pose two important questions regarding the attention mechanism.  In \autoref{att_feature}, we delve into a deep analysis of attention by answering the two proposed questions. Then, in \autoref{zoom}, we select a few counties as examples, magnifying their satellite images to demonstrate the role of attention. Furthermore, we summarize the advantages of attention in \autoref{att_over}. Finally, we discuss and explain the limitations of our method (\autoref{Limitation of our method}).

\subsection{Pixel mixing level calculation and preprocessing}
\label{corn_ratio}
To explore the role of attention, we first need to quantify the mixed pixels based on their definition, which is the proportion of the cornfield within a pixel of the feature dataset. Taking the MODIS dataset as an example, suppose one MODIS pixel contains $n$ CDL pixels. We represent each MODIS pixel as a binary list $S$ of length $n$, where each element $s_i$ represents a CDL pixel, $s_i \in \{ 0, 1\}$ for $i = 1,2,...,n$. Then, using GEE, we calculate the corn ratio, defined as the number of CDL pixels that represent corn within a MODIS pixel, to serve as the ground truth for our experiment:

\begin{align}
    corn \: ratio = \frac{\Sigma_{i=1}^n s_i}{n}
\end{align}

For the following analysis, the initial step of preprocessing is ``bag normalization'' by normalizing the corn ratio within each bag so that their sum equals one. Suppose we select $m$ pixels in each county $i$, $i=1,...,n$. In county $i$, we have a corn ratio $C_i^j$ for each pixel $j$, $j=1,...,m$. Then, the implementation of bag normalization is as follows:

\begin{align}
    {C_i^j}^\prime = \frac{C_i^j}{\Sigma_{k=1}^m C_i^k}
\end{align}

Where ${C_i^j}^\prime$ is bag-normalized corn ratio.

The purpose of bag normalization is to make attention and the corn ratio comparable. We first normalize the corn ratio values within each bag, transforming them into relative values. This ensures the fairness of subsequent experiments.

\subsection{Two questions towards attention mechanism}
\label{attention_analysis}
In the above experiments, we demonstrate that Att-MIL can indeed enhance the predictive ability of the model. To ascertain the specific role that attention plays within the model, further in-depth analysis is necessary. We need to answer two questions: (1) Which part of the input contributes the most? and (2) Do the weights learned by the attention mechanism truly reflect the degree of mixing in each pixel? In this section, we will demonstrate the specific function of attention by analyzing the relationship between attention, features, and corn ratio.  This experiment is based on the data from 2022, with the training data spanning from 2008 to 2021.

% \clearpage
\subsection{Attention’s correlation with features and corn ratio}
\label{att_feature}

\subsubsection{Feature importance}
\label{Feature importance}
To answer the first question, examining the importance of features is essential because we can verify whether the model derives attention based on relatively important features \citep{attentionsurvey}. To investigate feature importance, we employ the feature ablation strategy \citep{feature}. For each feature, we retrain our model after excluding that feature to test its impact on the model's performance (\autoref{tab:feature}). In this way, we can infer that if there's a notable decline in performance when a feature is removed, it indicates the feature's crucial significance.

\clearpage

\begin{table*}[ht]
\centering
\scalebox{0.75}{
\begin{tabular}{cccccccccc}
\toprule
\diagbox{Metrics}{Feature} & All & SOM & Year & CEC & PPT & VPDmin & VPDmax & Tmin & LSTnight\\
\midrule
$RMSE$ & 0.830 & 0.848 & 0.865 & 0.875 & 0.889 & 0.901 & 0.928 & 0.934 & 0.954 \\
\midrule
$R^2$ & 0.840 & 0.832 & 0.829 & 0.829 & 0.818 & 0.814 & 0.814 & 0.809 & 0.799 \\
\bottomrule
\end{tabular}
}
% \bigskip

\bigskip

\scalebox{0.75}{
\begin{tabular}{ccccccccc}
\toprule
\diagbox{Metrics}{Feature} & AWC & Tmean & NDWI & EVI & LSTday & GCI & Tmax & Historical \\
\midrule
$RMSE$ & 0.975 & 0.992 & 1.003 & 1.024 & 1.051 & 1.059 & 1.062 & 1.130 \\
\midrule
$R^2$ & 0.790 & 0.789 & 0.784 & 0.783 & 0.780 & 0.777 & 0.774 & 0.759 \\
\bottomrule
\end{tabular}
}
\caption{This table displays the results of our feature ablation experiments. All features are arranged in ascending order of importance, from left to right and from top to bottom.}
\label{tab:feature}
\end{table*}

% \clearpage

Through this examination, we found that the descending order of feature importance is as follows: historical average yield, Tmax, GCI, LSTday, EVI, NDWI, Tmean, AWC, LSTnight, Tmin, VPDmax, VPDmin, PPT, CEC, year, and SOM (\autoref{fig:feature}).  The historical average yield is considered to be of 100\% importance, and the least important feature SOM is treated as 0\% importance. All the features are arranged based on their experimental metrics. The importance of the other features is calculated using the formula:

\clearpage

\begin{figure*}[htbp]
\includegraphics[width=5.5in]{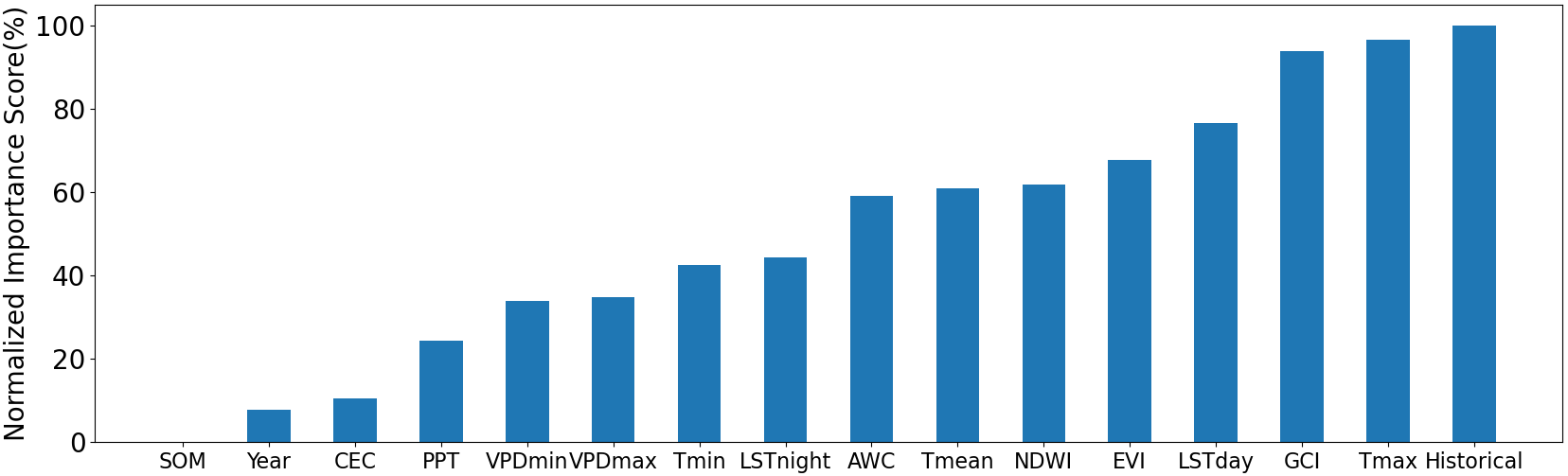}
\caption{The normalized importance ranking of a total of 16 features.}
\label{fig:feature}
\end{figure*}

% \clearpage

\begin{equation}
    \text{Importance} = \left( \frac{\text{value} - \text{min}}{\text{max} - \text{min}} \right) \times 100\%
\end{equation}

where $value$ is the metric ($RMSE$ and $R^2$) of the feature being evaluated, $min$ is the metric of the least important feature (SOM), and $max$ is the metric of the most important feature (historical average yield). This formula scales the importance of each feature relative to the range between the minimum and maximum metric values.

The most important feature, historical average yield, suggests that past yield data is crucial in predicting future yields. This aligns with the idea that past performance can often provide a strong baseline for future prediction. Tmax and GCI rank high, implying the importance of temperature and vegetation health in corn yield. On the other hand, the relatively low importance of features like year and SOM indicates that these features, while having some predictive power, contribute less significantly compared to other powerful features. We rank the importance of features here to delve deeper into their relationship with attention in subsequent analyses (\autoref{Featurecorr}).

\subsubsection{Attention's correlation with input features}
\label{Featurecorr}
 Since our model's input consists of various features, we aim to understand how the model learns attention from these features using the Pearson correlation coefficient \citep{pearson}. The Pearson correlation coefficient quantifies the linear relationship between two continuous variables. We identify some of the most important features in the feature importance experiment (\autoref{Feature importance}) and explore how the model learns attention by calculating the correlation coefficients between these features and attention. We perform the same pre-processing ``bag normalization'' as in \autoref{corn_ratio} to convert it into relative value.

As displayed in \autoref{tab:correlation}, the results indicate that attention has a high correlation with several important features, such as VIs, LSTday, and Tmax. In the context of statistical analysis, a correlation coefficient of 0.5 is significant enough to confirm that there is a strong correlation between the two factors \citep{correlation}. Since we know that the function of attention is to learn the part of the input that has a higher correlation with the output, we can conclude that the attention mechanism assigns attention scores by learning those important features.

\clearpage

\begin{table}[ht]
\centering
\scalebox{1}{
\begin{tabular}{ccccccc}
\toprule
Variables & NDWI & GCI & EVI & LSTday & Tmax & Corn ratio  \\
\midrule
Correlation & 0.50 & 0.53 & 0.48 & 0.45 & 0.52 & 0.68 \\

\bottomrule
\end{tabular}}
\caption{The Pearson correlation coefficient between attention with some of the most important features and corn ratio.}
\label{tab:correlation}
\end{table}

% \clearpage

\subsubsection{Attention's correlation with corn ratio}
\label{Attention's correlation with corn ratio}

To answer the second question, we explore whether the attention truly reflects the degree of mixing in mixed pixels, thereby assigning the correct weights to these pixels. Our goal is to empirically demonstrate the connection between the attention and the corn ratio. As displayed in \autoref{tab:correlation}, the results suggest a high correlation between the attention score and the corn ratio, indicating that our attention mechanism accurately predicts the degree of mixing in the mixed pixels.

\subsection{Zoomed-in mixed and pure pixels analysis}
\label{zoom}

\begin{figure}
    \captionsetup[subfigure]{labelformat=empty}
    \tikzset{inner sep=0pt}
    \setkeys{Gin}{width=0.22\textwidth}
    \centering
    
    \subfloat[\label{mysubfig1}]{%
    \tikz{\node (a) {\includegraphics[width=1.1in]{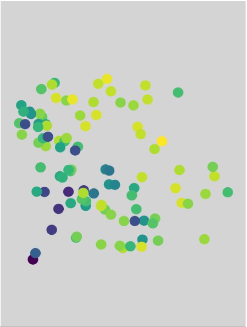}};
          \node[above, yshift=-0.5cm] at (a.south)  {(a1)};
          }}
    \hfil
    \subfloat[\label{mysubfig2}]{%
    \tikz{\node (a) {\includegraphics[width=1.1in]{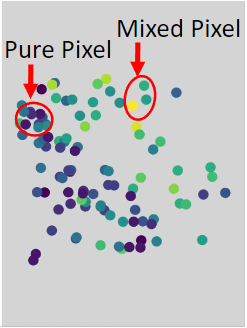}};
          \node[above, yshift=-0.5cm] at (a.south) {(b1)};
          }}
    \hfil
    \subfloat[\label{mysubfig3}]{%
    \tikz{\node (a) {\includegraphics[width=1.1in]{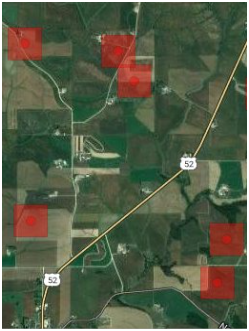}};
          \node[above, yshift=-0.5cm] at (a.south) {(c1)};
          }}
    \hfil
    \subfloat[\label{mysubfig4}]{%
    \tikz{\node (a) {\includegraphics[width=1.1in]{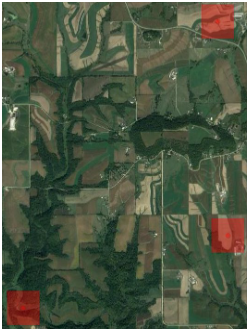}};
          \node[above, yshift=-0.5cm] at (a.south) {(d1)};
          }}

    \vspace{-0.5cm}
    
    \subfloat[\label{mysubfig5}]{%
    \tikz{\node (a) {\includegraphics[width=1.1in]{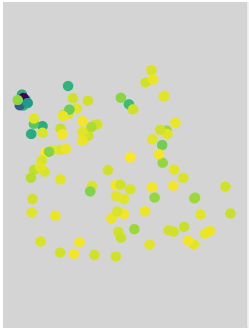}};
          \node[above, yshift=-0.5cm] at (a.south) {(a2)};
          }}
    \hfil
    \subfloat[\label{mysubfig6}]{%
    \tikz{\node (a) {\includegraphics[width=1.1in]{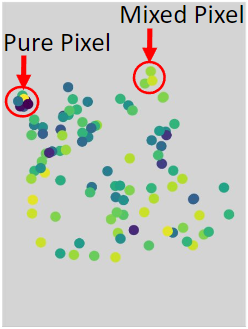}};
          \node[above, yshift=-0.5cm] at (a.south) {(b2)};
          }}
    \hfil
    \subfloat[\label{mysubfig7}]{%
    \tikz{\node (a) {\includegraphics[width=1.1in]{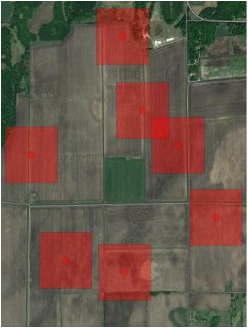}};
          \node[above, yshift=-0.5cm] at (a.south) {(c2)};
          }}
    \hfil
    \subfloat[\label{mysubfig8}]{%
    \tikz{\node (a) {\includegraphics[width=1.1in]{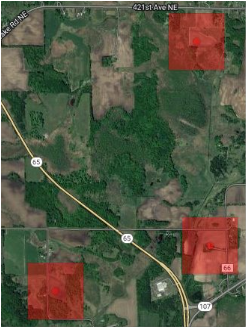}};
          \node[above, yshift=-0.5cm] at (a.south) {(d2)};
          }}

    \vspace{-0.5cm}
    
    \subfloat[\label{mysubfig9}]{%
    \tikz{\node (a) {\includegraphics[width=1.1in]{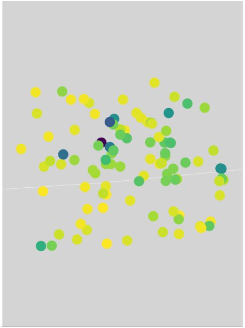}};
          \node[above, yshift=-0.5cm] at (a.south) {(a3)};
          }}
    \hfil
    \subfloat[\label{mysubfig10}]{%
    \tikz{\node (a) {\includegraphics[width=1.1in]{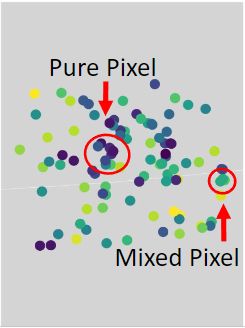}};
          \node[above, yshift=-0.5cm] at (a.south) {(b3)};
          }}
    \hfil
    \subfloat[\label{mysubfig11}]{%
    \tikz{\node (a) {\includegraphics[width=1.1in]{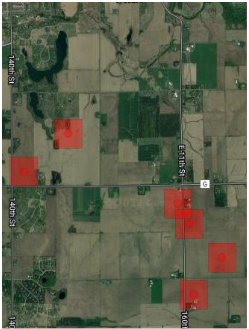}};
          \node[above, yshift=-0.5cm] at (a.south) {(c3)};
          }}
    \hfil
    \subfloat[\label{mysubfig12}]{%
    \tikz{\node (a) {\includegraphics[width=1.1in]{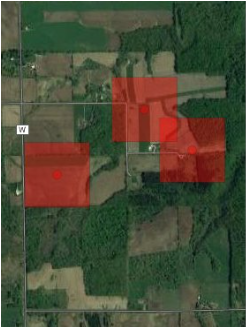}};
          \node[above, yshift=-0.5cm] at (a.south) {(d3)};
          }}

    \vspace{-0.5cm}
    
    \subfloat[\label{mysubfig13}]{%
    \tikz{\node (a) {\includegraphics[width=1.1in]{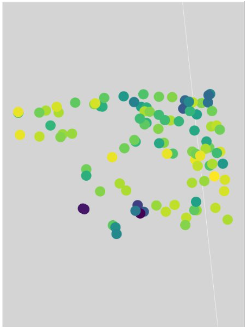}};
          \node[above, yshift=-0.5cm] at (a.south) {(a4)};
          }}
    \hfil
    \subfloat[\label{mysubfig14}]{%
    \tikz{\node (a) {\includegraphics[width=1.1in]{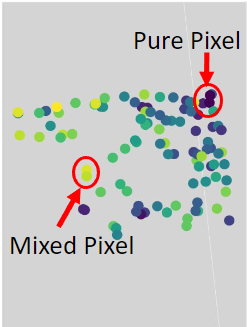}};
          \node[above, yshift=-0.5cm] at (a.south) {(b4)};
          }}
    \hfil
    \subfloat[\label{mysubfig15}]{%
    \tikz{\node (a) {\includegraphics[width=1.1in]{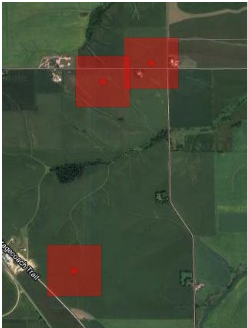}};
          \node[above, yshift=-0.5cm] at (a.south) {(c4)};
          }}
    \hfil
    \subfloat[\label{mysubfig16}]{%
    \tikz{\node (a) {\includegraphics[width=1.1in]{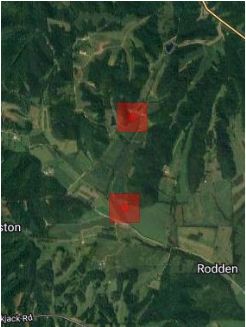}};
          \node[above, yshift=-0.5cm] at (a.south) {(d4)};
          }}
  \vspace{-0.5cm}
    \subfloat[\label{mysubfig16}]{%
    \tikz{\node (a) {\includegraphics[width=3in]{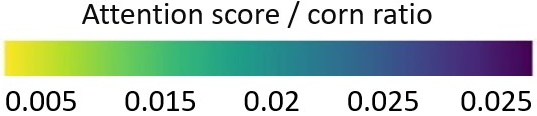}};
          }}
  \vspace{-0.7cm}
    \caption{Visual examination of (a) attention map, (b) corn ratio map, (c) zoomed-in pure pixels and (d) zoomed-in mixed pixels  in four areas (1-4) from top to bottom.}
    \label{fig:zoom}
\end{figure}

Previous experiments validated the roles of attention, corn ratio, and features from a statistical perspective. Based on our experimental setup, 100 CDL pixels are randomly selected in each county to represent the features of that county. We delve into each county to visually examine their connections (\autoref{fig:zoom}). Four counties are selected from all the data as examples from top to bottom. From left to right, the figures represent the attention map (a), corn ratio (b), a zoomed-in view of pure pixels (c), and a zoomed-in view of mixed pixels (d). In the visualization graphs of attention and corn ratio, the color of each point represents the magnitude of the value. The deeper the color, the larger the value. In the zoomed-in view of pixels, each small red dot represents the location of a CDL pixel. Furthermore, we aim to determine whether a CDL pixel is mixed or pure based on the surrounding land types. To this end, we plot $500\mathrm{m} \times 500\mathrm{m}$ red squares around the CDL pixels' locations to check whether the areas surrounding these CDL pixels contain cornfields or other types of land.

First, we examine the county-level attention visualization results from a broad perspective. By comparing the attention maps (\autoref{fig:zoom} a1-a4) with the corn ratio maps (\autoref{fig:zoom} b1-b4), we aim to demonstrate that our attention mechanism accurately reflects the proportion of cornfield pixels in each county. In \autoref{fig:zoom} (a1) and (b1), the top-left corner shows a cluster of darker-colored pure pixels, while the top-right corner features lighter-colored mixed pixels, indicating a strong correspondence between the attention and corn ratio. In \autoref{fig:zoom} (a2) and (b2), the left half of the pixels are darker, and the upper right section is lighter. Similarly, in \autoref{fig:zoom} (a3) and (b3), the center pixels are darker, while the right side pixels are lighter. In \autoref{fig:zoom} (a4) and (b4), the bottom left and top right sections of the pixels are darker, while the diagonal from the top left to the bottom right is lighter. These patterns clearly illustrate the correlation between our attention mechanism and the corn ratio at the county level.

\clearpage

% \clearpage

Next, we zoom into the clusters of pure pixels (c1-c4) and mixed pixels (d1-d4) to validate the above analysis. It is observed that in the pure pixels, all the land is cornfield, while in the mixed pixels, a significant portion of the land is occupied by unrelated categories such as forests. This fully demonstrates that our attention captures the information of the mixed cropland.

\subsection{Attention's advantage over corn ratio}
\label{att_over}

% It is observed that the correlation between attention and the corn ratio is not a perfect 100\%. The first reason is the inherent errors in the calculation of the corn ratio. This is attributed to the datasets' pixel sizes not being perfectly divisible and the boundaries of various pixels not matching seamlessly. As illustrated in \autoref{fig:mixedmodis}, the edges of the MODIS pixel do not align perfectly with the edges of the smaller CDL pixels. This results in many CDL pixels at the edges of MODIS pixels being only partially contained within the MODIS pixels. When we count these CDL pixels to calculate the corn ratio, it leads to inaccuracies. Therefore, we want the model to automatically learn attention, as we argue that the attention derived this way will be more accurate and reliable.

% The second reason is that our attention mechanism is a learned weight from all the features. In the above experiment, we calculate the corn ratio solely based on the MODIS dataset, which introduces certain errors. Therefore, our attention module provides a more comprehensive result compared to the corn ratio calculated solely from the MODIS dataset. 

A simple solution to the mixed pixel problem is to directly calculate the corn ratio for each mixed pixel, and use it to weight each pixel before predicting corn yield. We conducted an experiment here to demonstrate that this method is not a preferable solution. We calculate the corn ratio using the CDL mask and the MODIS dataset, and then explicitly apply this corn ratio as a weight to the features of each pixel. Suppose we select $100$ pixels in each county $i$, $i=1,...,n$. We have the corn ratio $C_i^j$, feature vector $P_i^j$, and corn yield $y_i$ for each pixel $j$, $j=1,...,100$. Furthermore, the implementation of manual weighting is as follows:

\begin{align}
    {P_i^j}^\prime = C_i^j P_i^j
\end{align}

Where ${P_i^j}^\prime$ is the manually-weighted feature vector.

Next, we use this manually weighted pixel dataset to predict corn yield $y_i$, namely $f({P^1_i}^\prime, {P^2_i}^\prime,..., {P^{100}_i}^\prime) = y_i$.  The model backbone is similar to that of our Att-MIL model. The only difference is the absence of the attention module; the manually-weighted method directly multiplies the corn ratio onto the pixel features instead. The experimental results are shown in \autoref{tab:manuallyweight}, which compares the results of manual weighting and automatic weighting (Auto-weighting) methods. We find that the results of using the corn ratio as weights are not ideal. This validates our argument that using attention to automatically learn the weights can achieve better results.

\clearpage

\begin{table}[ht]
\centering
\scalebox{1}{
\begin{tabular}{ccccc}
\toprule
\diagbox{Year}{Method} & \multicolumn{2}{c}{Auto-weighting} & \multicolumn{2}{c}{Manual weighting}  \\
\midrule
 & $RMSE$ & $R^2$& $RMSE$ & $R^2$\\
\hline
2018 & \textbf{1.00} & \textbf{0.67} & 1.35 & 0.47 \\
2019 & \textbf{0.99} & \textbf{0.58} & 1.27 & 0.55 \\
2020 & \textbf{1.00} & \textbf{0.48} & 1.43 & 0.31 \\
2021 & \textbf{0.99} & \textbf{0.64} & 1.24 & 0.57 \\
2022 & \textbf{0.83} & \textbf{0.84} & 1.17 & 0.69 \\
\bottomrule
\end{tabular}}
\caption{Comparison between manually calculating the mixing level of mixed pixels and applying weights versus automatically weighting with attention.}
\label{tab:manuallyweight}
\end{table}

% \clearpage

In addition to the significantly better experimental results as shown in \autoref{tab:manuallyweight}, the attention method also has other benefits over manual weighting. First, there are inherent errors in the calculation of the corn ratio. This is attributed to the datasets' pixel sizes not being perfectly divisible and the boundaries of various pixels not matching seamlessly. As illustrated in \autoref{fig:mixedmodis}, the boundaries of the MODIS pixels do not perfectly match those of the smaller CDL pixels. Consequently, many CDL pixels located at the boundaries of the MODIS pixels are only partially included within the MODIS pixels. When we count these CDL pixels to calculate the corn ratio, it leads to inaccuracies. Therefore, we use attention mechanism in our model to automatically learn attention, as the weights automatically derived by the attention method will yield more accurate and reliable predictions.

Second, our attention module can automatically learn a general weight for mixed pixels among different datasets. In \autoref{fig:mixedmodis} and the above experiments, we only use the MODIS dataset to illustrate the cause of mixed pixels and calculate the corn ratio. However, the problem of mixed pixels arises with any feature dataset with a resolution greater than $30\mathrm{m} \times 30\mathrm{m}$. When using many datasets with different resolutions, it would be necessary to calculate different corn ratios for each type of feature and then combine them for prediction, which is not a feasible or elegant solution. Our attention method does not require manually calculating the corn ratio for all datasets; instead, it automatically learns unified weights from different features. This method is more straightforward and elegant. These two reasons also help explain why the correlation between attention and the corn ratio is not a perfect 100\% (\autoref{Attention's correlation with corn ratio}), as our attention score is more accurate.

\subsection{Limitation of our method}
\label{Limitation of our method}

Although our Att-MIL method has been successful in pixel-level corn yield prediction and can effectively address the ``mixed pixel'' problem, there are still some limitations that need to be considered. First, due to the big size of satellite imagery, we can only extract VIs from it and use these as inputs for the model. This process results in some information loss, which is not entirely consistent with our original goal of pixel-level yield prediction. However, using the complete remote sensing images as input would demand substantial computational resources, making it challenging to balance these two aspects. Second, our method currently uses 16-day average features as input, which means that some subtle environmental changes cannot be captured. However, we consider that a half-month interval is sufficient to capture most relevant features when designing our experiment, and adding more data points would significantly increase the training cost of the model. Therefore, we believe using 16-day intervals is reasonable choice. Finally, while our model performs well in end-of-in-season prediction, it shows weaker performance in the early stages of in-season yield prediction. This reflects the shortcomings of learning-based methods, which require a large amount of data to achieve good predictive performance. In the future, we plan to incorporate process-based models to ensure better performance at earlier stages of corn growth.

% \clearpage
\section{Conclusion}
\label{Conclusion}

This paper explores county-level corn yield prediction by treating each county as a collection of cornfield pixels. To address the issues encountered in this specific scenario, we use a combination of MIL and attention mechanism as a solution. First, MIL is employed to leverage the pixel-level RS observations, balance the conflict between computational resources and information integrity, and address the lack of field-level yield records for pixel-level data processing. Subsequently, to solve the mixed pixel problem caused by inconsistent resolutions of feature datasets and crop mask, an attention mechanism is incorporated to assign weights to pixels, thereby enhancing prediction accuracy. We analyze the role of attention from both statistical and micro perspectives, confirming the validity of our approach. In the future, we will try to enhance the generalization of the model by transfer learning in larger-scale corn yield prediction research.

\section*{Acknowledgements}

We’d like to appreciate the editors and reviewers, whose comments are all valuable and very helpful for improving our manuscript.

\section*{Disclosure statement}
The authors report no conflict of interest.

\section*{Funding}

This work was supported by the United States Department of Agriculture (USDA) National Institute of Food and Agriculture, Agriculture and Food Research Initiative Project under Grant 1028199.

\clearpage

\bibliographystyle{tfcad}
\bibliography{interactcadsample}

\begin{thebibliography}{91}
\newcommand{\enquote}[1]{``#1''}
\providecommand{\natexlab}[1]{#1}
\providecommand{\url}[1]{\normalfont{#1}}
\providecommand{\urlprefix}{}

\bibitem[Albergel et~al.(2013)]{rmse}
Albergel, Clement, Luca Brocca, Wolfgang Wagner, Patricia de~Rosnay, and Jean-Christophe Calvet. 2013. ``Selection of performance metrics for global soil moisture products: The case of ASCAT product.'' \emph{Remote Sensing of Energy Fluxes and Soil Moisture Content} 427.

\bibitem[Bastiaanssen and Ali(2003)]{process1}
Bastiaanssen, Wim~GM, and Samia Ali. 2003. ``A new crop yield forecasting model based on satellite measurements applied across the Indus Basin, Pakistan.'' \emph{Agriculture, ecosystems \& environment} 94 (3): 321--340.

\bibitem[Battude et~al.(2016)]{rse6}
Battude, Marjorie, Ahmad Al~Bitar, David Morin, J{\'e}r{\^o}me Cros, Mireille Huc, Claire~Marais Sicre, Val{\'e}rie Le~Dantec, and Val{\'e}rie Demarez. 2016. ``Estimating maize biomass and yield over large areas using high spatial and temporal resolution Sentinel-2 like remote sensing data.'' \emph{Remote Sensing of Environment} 184: 668--681.

\bibitem[Bazrafshan et~al.(2022)]{mlp_2}
Bazrafshan, Ommolbanin, Mohammad Ehteram, Sarmad~Dashti Latif, Yuk~Feng Huang, Fang~Yenn Teo, Ali~Najah Ahmed, and Ahmed El-Shafie. 2022. ``Predicting crop yields using a new robust Bayesian averaging model based on multiple hybrid ANFIS and MLP models.'' \emph{Ain Shams Engineering Journal} 13 (5): 101724.

\bibitem[Bi et~al.(2023)]{transformer_3}
Bi, Luning, Owen Wally, Guiping Hu, Albert~U Tenuta, Yuba~R Kandel, and Daren~S Mueller. 2023. ``A transformer-based approach for early prediction of soybean yield using time-series images.'' \emph{Frontiers in Plant Science} 14: 1173036.

\bibitem[Bishop and Nasrabadi(2006)]{prml}
Bishop, Christopher~M, and Nasser~M Nasrabadi. 2006. \emph{Pattern recognition and machine learning}. Vol.~4. Springer.

\bibitem[Breiman(2001)]{rf}
Breiman, Leo. 2001. ``Random forests.'' \emph{Machine learning} 45: 5--32.

\bibitem[Carbonneau et~al.(2018)]{surveymir}
Carbonneau, Marc-Andr{\'e}, Veronika Cheplygina, Eric Granger, and Ghyslain Gagnon. 2018. ``Multiple instance learning: A survey of problem characteristics and applications.'' \emph{Pattern Recognition} 77: 329--353.

\bibitem[Daly et~al.(2008)]{prism2}
Daly, Christopher, Michael Halbleib, Joseph~I Smith, Wayne~P Gibson, Matthew~K Doggett, George~H Taylor, Jan Curtis, and Phillip~P Pasteris. 2008. ``Physiographically sensitive mapping of climatological temperature and precipitation across the conterminous United States.'' \emph{International Journal of Climatology: a Journal of the Royal Meteorological Society} 28 (15): 2031--2064.

\bibitem[Daly, Smith, and Olson(2015)]{prism1}
Daly, Christopher, Joseph~I Smith, and Keith~V Olson. 2015. ``Mapping atmospheric moisture climatologies across the conterminous United States.'' \emph{PloS one} 10 (10): e0141140.

\bibitem[Deines et~al.(2021)]{rse4}
Deines, Jillian~M, Rinkal Patel, Sang-Zi Liang, Walter Dado, and David~B Lobell. 2021. ``A million kernels of truth: Insights into scalable satellite maize yield mapping and yield gap analysis from an extensive ground dataset in the US Corn Belt.'' \emph{Remote sensing of environment} 253: 112174.

\bibitem[Dietterich, Lathrop, and Lozano-P{\'e}rez(1997)]{mir}
Dietterich, Thomas~G, Richard~H Lathrop, and Tom{\'a}s Lozano-P{\'e}rez. 1997. ``Solving the multiple instance problem with axis-parallel rectangles.'' \emph{Artificial intelligence} 89 (1-2): 31--71.

\bibitem[Dosovitskiy(2020)]{vit}
Dosovitskiy, Alexey. 2020. ``An image is worth 16x16 words: Transformers for image recognition at scale.'' \emph{arXiv preprint arXiv:2010.11929} .

\bibitem[Drought.gov(2018)]{kansas2018drought}
Drought.gov. 2018. ``Historical Drought Conditions in Kansas.'' \url{https://www.drought.gov/states/kansas}.

\bibitem[Drought.gov(2021{\natexlab{a}})]{nd2021drought}
Drought.gov. 2021{\natexlab{a}}. ``Historical Drought Conditions for North Dakota.'' \url{https://www.drought.gov/states/north-dakota}.

\bibitem[Drought.gov(2021{\natexlab{b}})]{sd2021drought}
Drought.gov. 2021{\natexlab{b}}. ``Historical Drought Conditions for South Dakota.'' \url{https://www.drought.gov/states/south-dakota}.

\bibitem[Drought.gov(2021{\natexlab{c}})]{mn2021drought}
Drought.gov. 2021{\natexlab{c}}. ``Historical Drought Conditions in Minnesota.'' \url{https://www.drought.gov/states/minnesota}.

\bibitem[Fan et~al.(2022)]{rnn_2}
Fan, Joshua, Junwen Bai, Zhiyun Li, Ariel Ortiz-Bobea, and Carla~P Gomes. 2022. ``A GNN-RNN approach for harnessing geospatial and temporal information: application to crop yield prediction.'' In \emph{Proceedings of the AAAI conference on artificial intelligence}, Vol.~36, 11873--11881.

\bibitem[Freedman, Pisani, and Purves(2007)]{pearson}
Freedman, David, Robert Pisani, and Roger Purves. 2007. ``Statistics (international student edition).'' \emph{Pisani, R. Purves, 4th edn. WW Norton \& Company, New York} .

\bibitem[Gao(1996)]{ndwi}
Gao, Bo-Cai. 1996. ``NDWI—A normalized difference water index for remote sensing of vegetation liquid water from space.'' \emph{Remote sensing of environment} 58 (3): 257--266.

\bibitem[Gitelson et~al.(2005)]{gci}
Gitelson, Anatoly~A, Andr{\'e}s Vi{\~n}a, Ver{\'o}nica Ciganda, Donald~C Rundquist, and Timothy~J Arkebauer. 2005. ``Remote estimation of canopy chlorophyll content in crops.'' \emph{Geophysical research letters} 32 (8).

\bibitem[Gong et~al.(2021)]{rnn_3}
Gong, Liyun, Miao Yu, Shouyong Jiang, Vassilis Cutsuridis, and Simon Pearson. 2021. ``Deep learning based prediction on greenhouse crop yield combined TCN and RNN.'' \emph{Sensors} 21 (13): 4537.

\bibitem[Goodfellow, Bengio, and Courville(2016)]{dlbook}
Goodfellow, Ian, Yoshua Bengio, and Aaron Courville. 2016. \emph{Deep learning}. MIT press.

\bibitem[Gorelick et~al.(2017)]{gee}
Gorelick, Noel, Matt Hancher, Mike Dixon, Simon Ilyushchenko, David Thau, and Rebecca Moore. 2017. ``Google Earth Engine: Planetary-scale geospatial analysis for everyone.'' \emph{Remote Sensing of Environment} https://doi.org/{10.1016/j.rse.2017.06.031},  \urlprefix\url{https://doi.org/10.1016/j.rse.2017.06.031}.

\bibitem[Haykin(1998)]{mlp}
Haykin, Simon. 1998. \emph{Neural networks: a comprehensive foundation}. Prentice Hall PTR.

\bibitem[He et~al.(2016)]{resnet}
He, Kaiming, Xiangyu Zhang, Shaoqing Ren, and Jian Sun. 2016. ``Deep residual learning for image recognition.'' In \emph{Proceedings of the IEEE conference on computer vision and pattern recognition}, Las Vegas, NV, USA, 770--778.

\bibitem[Hochreiter and Schmidhuber(1997)]{lstm}
Hochreiter, Sepp, and J{\"u}rgen Schmidhuber. 1997. ``Long short-term memory.'' \emph{Neural computation} 9 (8): 1735--1780.

\bibitem[Hoerl and Kennard(1970)]{ridge}
Hoerl, Arthur~E, and Robert~W Kennard. 1970. ``Ridge regression: Biased estimation for nonorthogonal problems.'' \emph{Technometrics} 12 (1): 55--67.

\bibitem[Hsieh, Lee, and Chen(2001)]{mix_2}
Hsieh, Pi-Fuei, Lou~C Lee, and Nai-Yu Chen. 2001. ``Effect of spatial resolution on classification errors of pure and mixed pixels in remote sensing.'' \emph{IEEE Transactions on Geoscience and Remote Sensing} 39 (12): 2657--2663.

\bibitem[Hu et~al.(2023)]{attmil2}
Hu, Huafeng, Ruijie Ye, Jeyan Thiyagalingam, Frans Coenen, and Jionglong Su. 2023. ``Triple-kernel gated attention-based multiple instance learning with contrastive learning for medical image analysis.'' \emph{Applied Intelligence} 1--16.

\bibitem[Huang et~al.(2015)]{process2}
Huang, Jianxi, Hongyuan Ma, Wei Su, Xiaodong Zhang, Yanbo Huang, Jinlong Fan, and Wenbin Wu. 2015. ``Jointly assimilating MODIS LAI and ET products into the SWAP model for winter wheat yield estimation.'' \emph{IEEE Journal of Selected Topics in Applied Earth Observations and Remote Sensing} 8 (8): 4060--4071.

\bibitem[Huete et~al.(2002)]{evi}
Huete, Alfredo, Kamel Didan, Tomoaki Miura, E~Patricia Rodriguez, Xiang Gao, and Laerte~G Ferreira. 2002. ``Overview of the radiometric and biophysical performance of the MODIS vegetation indices.'' \emph{Remote sensing of environment} 83 (1-2): 195--213.

\bibitem[Ilse, Tomczak, and Welling(2018)]{attmir}
Ilse, Maximilian, Jakub Tomczak, and Max Welling. 2018. ``Attention-based deep multiple instance learning.'' In \emph{International conference on machine learning}, 2127--2136. PMLR.

\bibitem[InForum(2021)]{nd2021temp}
InForum. 2021. ``2021 was North Dakota's 5th warmest year on record.'' \url{https://www.inforum.com/news/north-dakota/2021-was-north-dakotas-5th-warmest-year-on-record}.

\bibitem[Jiang et~al.(2020)]{lstm_1}
Jiang, Hao, Hao Hu, Renhai Zhong, Jinfan Xu, Jialu Xu, Jingfeng Huang, Shaowen Wang, Yibin Ying, and Tao Lin. 2020. ``A deep learning approach to conflating heterogeneous geospatial data for corn yield estimation: A case study of the US Corn Belt at the county level.'' \emph{Global change biology} 26 (3): 1754--1766.

\bibitem[Jin et~al.(2019)]{rse5}
Jin, Zhenong, George Azzari, Calum You, Stefania Di~Tommaso, Stephen Aston, Marshall Burke, and David~B Lobell. 2019. ``Smallholder maize area and yield mapping at national scales with Google Earth Engine.'' \emph{Remote Sensing of Environment} 228: 115--128.

\bibitem[Johnson(2014{\natexlab{a}})]{rse8}
Johnson, David~M. 2014{\natexlab{a}}. ``An assessment of pre-and within-season remotely sensed variables for forecasting corn and soybean yields in the United States.'' \emph{Remote Sensing of Environment} 141: 116--128.

\bibitem[Johnson(2014{\natexlab{b}})]{inseason}
Johnson, David~M. 2014{\natexlab{b}}. ``An assessment of pre-and within-season remotely sensed variables for forecasting corn and soybean yields in the United States.'' \emph{Remote Sensing of Environment} 141: 116--128.

\bibitem[Justice et~al.(1998)]{modis}
Justice, Christopher~O, Eric Vermote, John~RG Townshend, Ruth Defries, David~P Roy, Dorothy~K Hall, Vincent~V Salomonson, et~al. 1998. ``The Moderate Resolution Imaging Spectroradiometer (MODIS): Land remote sensing for global change research.'' \emph{IEEE transactions on geoscience and remote sensing} 36 (4): 1228--1249.

\bibitem[Khaki and Wang(2019)]{mlp_1}
Khaki, Saeed, and Lizhi Wang. 2019. ``Crop yield prediction using deep neural networks.'' \emph{Frontiers in plant science} 10: 621.

\bibitem[Khaki, Wang, and Archontoulis(2020{\natexlab{a}})]{cnn_2}
Khaki, Saeed, Lizhi Wang, and Sotirios~V Archontoulis. 2020{\natexlab{a}}. ``A CNN-RNN framework for crop yield prediction.'' \emph{Frontiers in Plant Science} 10: 1750.

\bibitem[Khaki, Wang, and Archontoulis(2020{\natexlab{b}})]{rnn_1}
Khaki, Saeed, Lizhi Wang, and Sotirios~V Archontoulis. 2020{\natexlab{b}}. ``A cnn-rnn framework for crop yield prediction.'' \emph{Frontiers in Plant Science} 10: 1750.

\bibitem[Khalifani et~al.(2022)]{mlp_3}
Khalifani, Sanaz, Reza Darvishzadeh, Nasrin Azad, and Razgar~Seyed Rahmani. 2022. ``Prediction of sunflower grain yield under normal and salinity stress by RBF, MLP and, CNN models.'' \emph{Industrial Crops and Products} 189: 115762.

\bibitem[Khan et~al.(2022)]{attentionsurvey}
Khan, Salman, Muzammal Naseer, Munawar Hayat, Syed~Waqas Zamir, Fahad~Shahbaz Khan, and Mubarak Shah. 2022. ``Transformers in vision: A survey.'' \emph{ACM computing surveys (CSUR)} 54 (10s): 1--41.

\bibitem[Kingma and Ba(2014)]{adam}
Kingma, Diederik~P, and Jimmy Ba. 2014. ``Adam: A method for stochastic optimization.'' \emph{arXiv preprint arXiv:1412.6980} .

\bibitem[Krishnan et~al.(2024)]{transformer_4}
Krishnan, V~Gokula, BV~Subba Rao, J~Rajendra Prasad, P~Pushpa, and S~Kumari. 2024. ``Sugarcane yield prediction using NOA-based swin transformer model in IoT smart agriculture.'' \emph{Journal of Applied Biology \& Biotechnology Vol} 12 (2): 239--247.

\bibitem[LeCun, Bengio et~al.(1995)]{cnn}
LeCun, Yann, Yoshua Bengio, et~al. 1995. ``Convolutional networks for images, speech, and time series.'' \emph{The handbook of brain theory and neural networks} 3361 (10): 1995.

\bibitem[Lin et~al.(2023)]{transformer_2}
Lin, Fudong, Summer Crawford, Kaleb Guillot, Yihe Zhang, Yan Chen, Xu~Yuan, Li~Chen, et~al. 2023. ``MMST-ViT: Climate Change-aware Crop Yield Prediction via Multi-Modal Spatial-Temporal Vision Transformer.'' In \emph{Proceedings of the IEEE/CVF International Conference on Computer Vision}, 5774--5784.

\bibitem[Litkowski(2016)]{feature}
Litkowski, Ken. 2016. ``Feature ablation for preposition disambiguation.'' \emph{Damascus, MD, USA: CL Research} .

\bibitem[Liu et~al.(2022{\natexlab{a}})]{transformer_0}
Liu, Yuanyuan, Shaoqiang Wang, Jinghua Chen, Bin Chen, Xiaobo Wang, Dongze Hao, and Leigang Sun. 2022{\natexlab{a}}. ``Rice yield prediction and model interpretation based on satellite and climatic indicators using a transformer method.'' \emph{Remote Sensing} 14 (19): 5045.

\bibitem[Liu et~al.(2022{\natexlab{b}})]{transformer_1}
Liu, Yuanyuan, Shaoqiang Wang, Jinghua Chen, Bin Chen, Xiaobo Wang, Dongze Hao, and Leigang Sun. 2022{\natexlab{b}}. ``Rice yield prediction and model interpretation based on satellite and climatic indicators using a transformer method.'' \emph{Remote Sensing} 14 (19): 5045.

\bibitem[Liu et~al.(2021)]{swin}
Liu, Ze, Yutong Lin, Yue Cao, Han Hu, Yixuan Wei, Zheng Zhang, Stephen Lin, and Baining Guo. 2021. ``Swin transformer: Hierarchical vision transformer using shifted windows.'' In \emph{Proceedings of the IEEE/CVF international conference on computer vision}, 10012--10022.

\bibitem[Lobell and Burke(2010)]{process4}
Lobell, David~B, and Marshall~B Burke. 2010. ``On the use of statistical models to predict crop yield responses to climate change.'' \emph{Agricultural and forest meteorology} 150 (11): 1443--1452.

\bibitem[Ma et~al.(2024)]{yuchi8}
Ma, Yuchi, Shuo Chen, Stefano Ermon, and David~B. Lobell. 2024. ``Transfer learning in environmental remote sensing.'' \emph{Remote Sensing of Environment} 301: 113924. https://doi.org/{https://doi.org/10.1016/j.rse.2023.113924},  \urlprefix\url{https://www.sciencedirect.com/science/article/pii/S0034425723004765}.

\bibitem[Ma et~al.(2019)]{yuchi6}
Ma, Yuchi, Yanghui Kang, Mutlu Ozdogan, and Zhou Zhang. 2019. ``County-level corn yield prediction using deep transfer learning.'' In \emph{AGU Fall Meeting Abstracts}, Vol. 2019, B54D--02.

\bibitem[Ma et~al.(2023)]{yuchi7}
Ma, Yuchi, Zhengwei Yang, Qunying Huang, and Zhou Zhang. 2023. ``Improving the Transferability of Deep Learning Models for Crop Yield Prediction: A Partial Domain Adaptation Approach.'' \emph{Remote Sensing} 15 (18): 4562.

\bibitem[Ma, Yang, and Zhang(2023)]{yuchi1}
Ma, Yuchi, Zhengwei Yang, and Zhou Zhang. 2023. ``Multisource Maximum Predictor Discrepancy for Unsupervised Domain Adaptation on Corn Yield Prediction.'' \emph{IEEE Transactions on Geoscience and Remote Sensing} 61: 1--15.

\bibitem[Ma and Zhang(2022)]{yuchi4}
Ma, Yuchi, and Zhou Zhang. 2022. ``Multi-source unsupervised domain adaptation on corn yield prediction.'' In \emph{AI for Agriculture and Food Systems}, .

\bibitem[Ma et~al.(2021{\natexlab{a}})]{rse3}
Ma, Yuchi, Zhou Zhang, Yanghui Kang, and Mutlu {\"O}zdo{\u{g}}an. 2021{\natexlab{a}}. ``Corn yield prediction and uncertainty analysis based on remotely sensed variables using a Bayesian neural network approach.'' \emph{Remote Sensing of Environment} 259: 112408.

\bibitem[Ma et~al.(2021{\natexlab{b}})]{yuchi5}
Ma, Yuchi, Zhou Zhang, Hsiuhan~Lexie Yang, and Zhengwei Yang. 2021{\natexlab{b}}. ``An adaptive adversarial domain adaptation approach for corn yield prediction.'' \emph{Computers and Electronics in Agriculture} 187: 106314.

\bibitem[Mishra et~al.(2013)]{process3}
Mishra, Vikalp, James~F Cruise, John~R Mecikalski, Christopher~R Hain, and Martha~C Anderson. 2013. ``A remote-sensing driven tool for estimating crop stress and yields.'' \emph{Remote Sensing} 5 (7): 3331--3356.

\bibitem[NASS(2018)]{kansas2018yield}
NASS, USDA. 2018. ``USDA NASS 2018 Kansas Corn Statistics.'' \url{https://kscorn.com/2019/02/08/usda-nass-2018-kansas-corn-statistics/}.

\bibitem[Nevavuori, Narra, and Lipping(2019)]{cnn_1}
Nevavuori, Petteri, Nathaniel Narra, and Tarmo Lipping. 2019. ``Crop yield prediction with deep convolutional neural networks.'' \emph{Computers and electronics in agriculture} 163: 104859.

\bibitem[of~Natural~Resources(2021)]{mn2021drought1}
of~Natural~Resources, Minnesota~Department. 2021. ``The Drought of 2021.'' \url{https://www.dnr.state.mn.us/climate/journal/drought-2021.html}.

\bibitem[Park, Feddema, and Egbert(2005)]{modismyd}
Park, S, JJ~Feddema, and SL~Egbert. 2005. ``MODIS land surface temperature composite data and their relationships with climatic water budget factors in the central Great Plains.'' \emph{International Journal of Remote Sensing} 26 (6): 1127--1144.

\bibitem[Paszke et~al.(2019)]{pytorch}
Paszke, Adam, Sam Gross, Francisco Massa, Adam Lerer, James Bradbury, Gregory Chanan, Trevor Killeen, et~al. 2019. ``Pytorch: An imperative style, high-performance deep learning library.'' \emph{Advances in neural information processing systems} 32.

\bibitem[Pedregosa et~al.(2011)]{sklearn}
Pedregosa, F., G.~Varoquaux, A.~Gramfort, V.~Michel, B.~Thirion, O.~Grisel, M.~Blondel, et~al. 2011. ``Scikit-learn: Machine Learning in {P}ython.'' \emph{Journal of Machine Learning Research} 12: 2825--2830.

\bibitem[Qi et~al.(2017)]{qi2017pointnet}
Qi, Charles~R, Hao Su, Kaichun Mo, and Leonidas~J Guibas. 2017. ``Pointnet: Deep learning on point sets for 3d classification and segmentation.'' In \emph{Proceedings of the IEEE conference on computer vision and pattern recognition}, 652--660.

\bibitem[Ray and Craven(2005)]{insmir}
Ray, Soumya, and Mark Craven. 2005. ``Supervised versus multiple instance learning: An empirical comparison.'' In \emph{Proceedings of the 22nd international conference on Machine learning}, 697--704.

\bibitem[Ray and Page(2001)]{primemir}
Ray, Soumya, and David Page. 2001. ``Multiple Instance Regression.'' In \emph{Proceedings of the Eighteenth International Conference on Machine Learning {(ICML} 2001), Williams College, Williamstown, MA, USA, June 28 - July 1, 2001},  edited by Carla~E. Brodley and Andrea~Pohoreckyj Danyluk, 425--432. Morgan Kaufmann.

\bibitem[Rumelhart et~al.(1985)]{rnn}
Rumelhart, David~E, Geoffrey~E Hinton, Ronald~J Williams, et~al. 1985. ``Learning internal representations by error propagation.'' .

\bibitem[Rymarczyk et~al.(2021)]{attmil1}
Rymarczyk, Dawid, Adriana Borowa, Jacek Tabor, and Bartosz Zielinski. 2021. ``Kernel self-attention for weakly-supervised image classification using deep multiple instance learning.'' In \emph{Proceedings of the IEEE/CVF Winter Conference on Applications of Computer Vision}, 1721--1730.

\bibitem[Sagan et~al.(2021)]{resnet_1}
Sagan, Vasit, Maitiniyazi Maimaitijiang, Sourav Bhadra, Matthew Maimaitiyiming, Davis~R Brown, Paheding Sidike, and Felix~B Fritschi. 2021. ``Field-scale crop yield prediction using multi-temporal WorldView-3 and PlanetScope satellite data and deep learning.'' \emph{ISPRS journal of photogrammetry and remote sensing} 174: 265--281.

\bibitem[Sakamoto, Gitelson, and Arkebauer(2013)]{rse9}
Sakamoto, Toshihiro, Anatoly~A Gitelson, and Timothy~J Arkebauer. 2013. ``MODIS-based corn grain yield estimation model incorporating crop phenology information.'' \emph{Remote Sensing of Environment} 131: 215--231.

\bibitem[Sakamoto, Gitelson, and Arkebauer(2014)]{rse7}
Sakamoto, Toshihiro, Anatoly~A Gitelson, and Timothy~J Arkebauer. 2014. ``Near real-time prediction of US corn yields based on time-series MODIS data.'' \emph{Remote Sensing of Environment} 147: 219--231.

\bibitem[Schaaf and Wang(2015)]{modismcd}
Schaaf, C, and Z~Wang. 2015. ``MCD43A4 MODIS/Terra+ Aqua BRDF/Albedo Nadir BRDF Adjusted Ref Daily L3 Global-500m V006. NASA EOSDIS Land Processes DAAC.'' \emph{USGS Earth Resources Observation and Science (EROS) Center, Sioux Falls, South Dakota (https://lpdaac. usgs. gov)} .

\bibitem[Schober, Boer, and Schwarte(2018)]{correlation}
Schober, Patrick, Christa Boer, and Lothar~A Schwarte. 2018. ``Correlation coefficients: appropriate use and interpretation.'' \emph{Anesthesia \& analgesia} 126 (5): 1763--1768.

\bibitem[Service(2020)]{iowa2020stormnws}
Service, National~Weather. 2020. ``August 10, 2020 Derecho.'' \url{https://www.weather.gov/dmx/2020derecho}.

\bibitem[Shahhosseini et~al.(2021{\natexlab{a}})]{county1}
Shahhosseini, Mohsen, Guiping Hu, Isaiah Huber, and Sotirios~V Archontoulis. 2021{\natexlab{a}}. ``Coupling machine learning and crop modeling improves crop yield prediction in the US Corn Belt.'' \emph{Scientific reports} 11 (1): 1606.

\bibitem[Shahhosseini et~al.(2021{\natexlab{b}})]{cnn_3}
Shahhosseini, Mohsen, Guiping Hu, Saeed Khaki, and Sotirios~V Archontoulis. 2021{\natexlab{b}}. ``Corn yield prediction with ensemble CNN-DNN.'' \emph{Frontiers in plant science} 12: 709008.

\bibitem[Shuai and Basso(2022)]{rse1}
Shuai, Guanyuan, and Bruno Basso. 2022. ``Subfield maize yield prediction improves when in-season crop water deficit is included in remote sensing imagery-based models.'' \emph{Remote Sensing of Environment} 272: 112938.

\bibitem[Sun et~al.(2019)]{cnn_lstm_1}
Sun, Jie, Liping Di, Ziheng Sun, Yonglin Shen, and Zulong Lai. 2019. ``County-level soybean yield prediction using deep CNN-LSTM model.'' \emph{Sensors} 19 (20): 4363.

\bibitem[USDA(2020{\natexlab{a}})]{ssurgo}
USDA. 2020{\natexlab{a}}. ``Soil Survey Staff, Natural Resources Conservation Service, United States Department of Agriculture. Web Soil Survey. Available online at https://websoilsurvey.nrcs.usda.gov/. Accessed [12/06/2020].''  .

\bibitem[USDA(2020{\natexlab{b}})]{yielddata}
USDA. 2020{\natexlab{b}}. ``United States Department of Agriculture National Agricultural Statistics Service.''  .

\bibitem[USDA-NASS(2017)]{cdl}
USDA-NASS, CDL. 2017. ``USDA national agricultural statistics service cropland data layer.''  .

\bibitem[Vaswani et~al.(2017)]{attention}
Vaswani, Ashish, Noam Shazeer, Niki Parmar, Jakob Uszkoreit, Llion Jones, Aidan~N Gomez, {\L}ukasz Kaiser, and Illia Polosukhin. 2017. ``Attention is all you need.'' \emph{Advances in neural information processing systems} 30.

\bibitem[Wagstaff, Lane, and Roper(2008)]{clustermir}
Wagstaff, Kiri~L, Terran Lane, and Alex Roper. 2008. ``Multiple-instance regression with structured data.'' In \emph{2008 IEEE international conference on data mining workshops}, 291--300. IEEE.

\bibitem[Wang et~al.(2020)]{county2}
Wang, Xinlei, Jianxi Huang, Quanlong Feng, and Dongqin Yin. 2020. ``Winter wheat yield prediction at county level and uncertainty analysis in main wheat-producing regions of China with deep learning approaches.'' \emph{Remote Sensing} 12 (11): 1744.

\bibitem[Xu et~al.(2005)]{mix_1}
Xu, Min, Pakorn Watanachaturaporn, Pramod~K Varshney, and Manoj~K Arora. 2005. ``Decision tree regression for soft classification of remote sensing data.'' \emph{Remote Sensing of Environment} 97 (3): 322--336.

\bibitem[Yang et~al.(2021)]{rse2}
Yang, Yang, Martha~C Anderson, Feng Gao, David~M Johnson, Yun Yang, Liang Sun, Wayne Dulaney, et~al. 2021. ``Phenological corrections to a field-scale, ET-based crop stress indicator: An application to yield forecasting across the US Corn Belt.'' \emph{Remote Sensing of Environment} 257: 112337.

\bibitem[Zaheer et~al.(2017)]{zaheer2017deep}
Zaheer, Manzil, Satwik Kottur, Siamak Ravanbakhsh, Barnabas Poczos, Russ~R Salakhutdinov, and Alexander~J Smola. 2017. ``Deep sets.'' \emph{Advances in neural information processing systems} 30.

\end{thebibliography}

\end{document}